\documentclass{article} 
\usepackage[final]{colm2026_conference}

\usepackage{microtype}
\usepackage{hyperref}
\usepackage{url}
\usepackage{booktabs}

\usepackage{lineno}

\usepackage[T1]{fontenc}
\usepackage{multirow}
\usepackage{fdsymbol}
\usepackage[most]{tcolorbox}
\usepackage{kotex}
\usepackage{xspace}
\usepackage{enumitem}
\usepackage{pifont}
\usepackage[dvipsnames]{xcolor}
\usepackage[normalem]{ulem}
\usepackage{expex}
\usepackage{subcaption}
\usepackage{wrapfig}
\usepackage[export]{adjustbox}
\usepackage{fontawesome}
\def\eg{\emph{e.g}.,\xspace}
\def\ie{\emph{i.e}.,\xspace}
\def\method{\textbf{\texttt{CSICL}}\xspace}
\newcommand{\cmark}{\textcolor{Green}{\ding{51}}}
\newcommand{\xmark}{\ding{55}}
\useunder{\uline}{\ul}{}

\definecolor{darkblue}{rgb}{0, 0, 0.5}
\hypersetup{colorlinks=true, citecolor=darkblue, linkcolor=darkblue, urlcolor=darkblue}

\title{Gradual Code-Switching as Inference-Time \\Cross-Lingual Representational Alignment for LLMs}

\author{Haneul Yoo$^\heartsuit$ \hspace{1em} Jiho Jin$^\clubsuit$ \hspace{1em} Kyunghyun Cho$^\heartsuit$ \hspace{1em} Alice Oh$^\clubsuit$ \\
$^\heartsuit$New York University~~~$^\clubsuit$KAIST \\
\texttt{haneul.yoo@nyu.edu}
}
\begin{document}

\ifcolmsubmission
\linenumbers
\fi

\maketitle

\begin{abstract}
While large language models (LLMs) have achieved notable progress in multilingual settings, their performance remains uneven across languages as LLMs often rely on English-centric latent representations.
In this work, we introduce code-switching in-context learning (\method), an inference-time mechanism for cross-lingual representational alignment.
Rather than relying on an abrupt translation pivot, \method explicitly scaffolds the reasoning trajectory by gradually transitioning from a target language to English, aligning non-English inputs with an English-centric reasoning space.
Across 4 LLMs, 6 datasets, and 10 languages, \method consistently outperforms cross-lingual in-context learning baselines, yielding average gains of 6.0pp and 4.8pp in target and unseen languages, respectively.
The improvements generalize across language families and are even more pronounced in low-resource settings, with gains of 14.7pp in target and 5.3pp in unseen languages.
These findings establish code-switching as a robust and effective approach for reducing cross-lingual misalignment during inference, moving LLMs towards more equitable and effective multilingual systems.

{\centering \faGithub~\url{https://github.com/haneul-yoo/csicl/} \par}
\end{abstract}

\begin{figure}[ht]
    \centering
    \includegraphics[width=.6\linewidth]{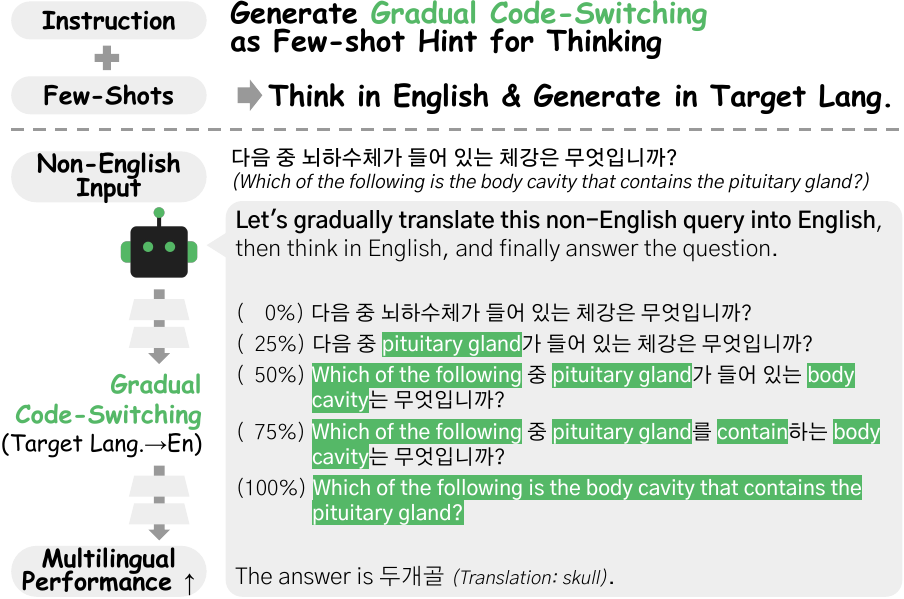}
    \caption{Overview of \method, which gradually transitions from a target language to English to better align cross-lingual representations.}
    \label{fig:teaser_image}
\end{figure}

\section{Introduction}
Large language models (LLMs) have demonstrated remarkable multilingual capabilities, powering diverse tasks such as question answering~\citep{tjuatja2024do}, translation~\citep{xu2024a}, and reasoning~\citep{shi2023language}.
However, recent studies reveal that this competence is far from language-agnostic, as LLMs typically rely on English-centric latent representations~\citep{wendler-etal-2024-llamas, zhong-etal-2025-language}.
Consequently, cross-lingual misalignment, such as failures in internal translations into English, can lead to a sharp drop in non-English performance~\citep{bafna2025translation}.
This highlights that multilingual competence in LLMs remains uneven, thereby limiting the inclusive deployment of these models across diverse linguistic communities around the world.

A common approach to improving multilingual performance is cross-lingual in-context learning (X-ICL), which uses demonstrations in high-resource languages (\eg English)~\citep{lin-etal-2022-shot} to elicit task behavior in low-resource~\citep{cahyawijaya-etal-2024-llms} or unseen~\citep{winata-etal-2022-cross} languages.
However, existing X-ICL often transfers superficial task patterns (\eg output format, label mapping, etc.), employing monolingual demonstrations as an abrupt single-step pivot without necessarily reducing the underlying cross-lingual representational gap.

Meanwhile, code-switching, a natural alternation between languages, has been shown to benefit cross-lingual transfer in training phases~\citep{wang-etal-2025-investigating-scaling, yoo-etal-2025-code-switching, chai2025xcot}, pointing to its potential to improve cross-lingual alignment at inference time.
In this paper, we propose code-switching in-context learning (\method), a novel inference-time representational alignment mechanism.
Instead of relying on an abrupt single-step transition pivot, \method explicitly scaffolds the reasoning trajectory of LLMs through a gradual transition from a target language to English during inference (Figure~\ref{fig:teaser_image}).
Specifically, \method begins in the target language, progressively introduces English tokens via code-switching, and converges to a full English equivalent.
This gradual shift acts as a \emph{linguistic bridge}, nudging LLMs to dynamically align cross-lingual representations directly, rather than relying solely on an abrupt latent translation.

We evaluate \method across 4 multilingual LLMs in 10 languages on 6 datasets spanning knowledge-intensive and reasoning-oriented tasks.
We observe that \method consistently outperforms X-ICL baselines, yielding improvements of 6.0pp and 4.8pp in target and unseen languages, respectively.
We generalize this to diverse tasks and highlight its effectiveness, particularly in translation and reasoning tasks.
These findings position \method as a robust, effective mechanism for mitigating cross-lingual misalignment that controls gradual cross-lingual transitions at inference time.

Our contributions are as follows:
\begin{itemize}[leftmargin=*, noitemsep]
    \item We propose \method, an inference-time cross-lingual representation alignment mechanism that gradually bridges target languages with English representations during inference.
    \item We provide systematic empirical evidence across 4 LLMs, 6 tasks, and 10 languages that \method consistently improves cross-lingual transfer over standard X-ICL baselines, especially in unseen and low-resource languages.
    \item We visualize hidden states where \method systematically aligns non-English inputs with the English-centric latent space. We further analyze key design factors of \method, including granularity, directionality, pivot language, and generation strategy.
\end{itemize}

\section{Code-Switching In-Context Learning}
We propose \method, a training-free inference-time method for improving cross-lingual transfer of LLMs.
Figure~\ref{fig:teaser_image} illustrates an abstract example of \method.
Given (1) a gradual translation instruction and (2) gradual code-switching few-shot demonstrations, \method constructs a stepwise transition from a target language to English.

\subsection{Inference-Time Transition via Gradual Code-Switching}

\paragraph{Instruction.}
We instruct LLMs to process a non-English input through a gradual transition into English.
Specifically, we ask them to 
\begin{enumerate}[leftmargin=*, noitemsep]
    \item Repeat a short, single-sentence instruction that guides a model to gradual translation (\ie ``\emph{Let's gradually translate this non-English query into English, then think in English, and finally answer the question.''});
    \item Explicitly show their step-by-step, progressive translation process into English;
    \item Provide the final answer in the target language, following the specified format (\eg ``\emph{The answer is \ldots}'').
\end{enumerate}

\paragraph{Demonstrations.}
In addition to the instruction, \method provides few-shot demonstrations that instantiate the same transition pattern, where each demonstration:
\begin{enumerate}[leftmargin=*, noitemsep]
    \item Begins with a query in a target language (\ie En 0\%);
    \item Progressively transitions to English by leveraging code-switching whose matrix language is the target language and the embedded language is English (\ie En 25\textrightarrow 50\textrightarrow 75\%);
    \item Finally concludes with the full English equivalent (\ie En 100\%).
\end{enumerate}

\subsection{Demonstration Construction}

\begin{figure}[h]
    \centering
    \includegraphics[width=0.75\linewidth]{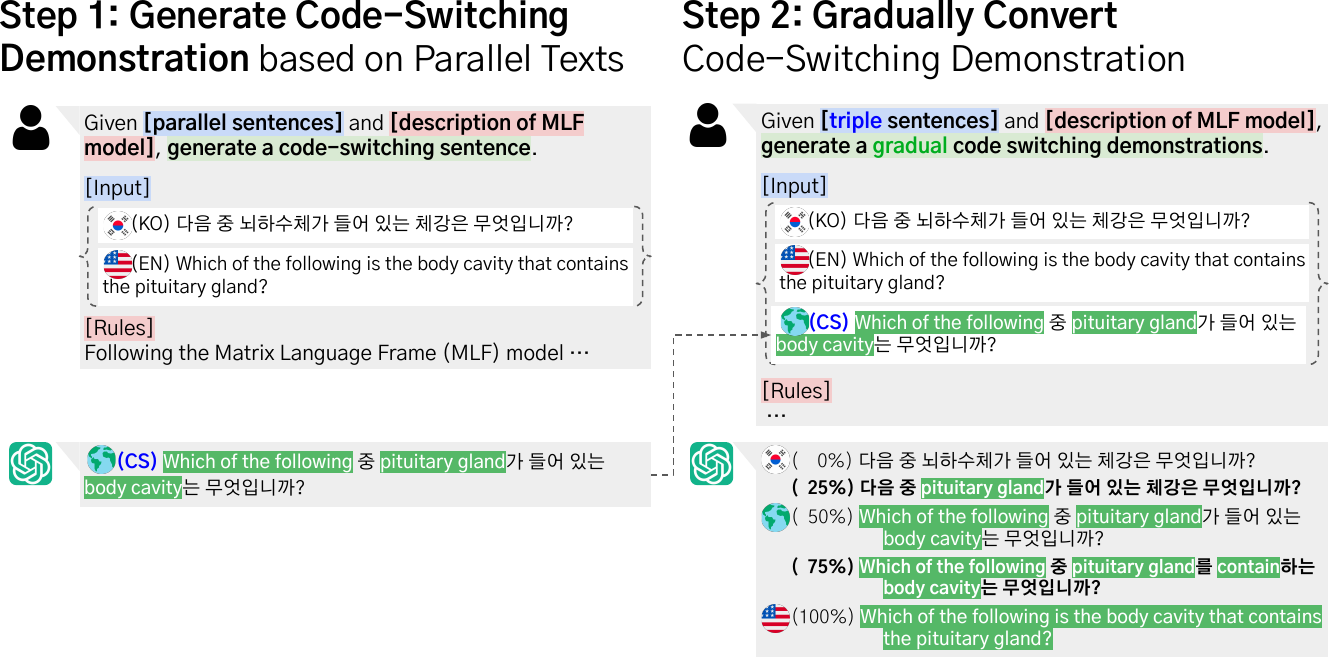}
    \caption{Two-step pipeline to generate gradual code-switching few-shot demonstrations.}
    \label{fig:demonstration_generation}
\end{figure}

For each target language, we randomly sample 5 instances as demonstrations and convert them into gradual code-switching sequences.
Figure~\ref{fig:demonstration_generation} illustrates the two-step pipeline to construct these demonstrations in \method.
Following \citet{kim2025codeswitched}, we first generate the code-switching sentences given parallel inputs, and then construct a gradual code-switching sequence increasing the English proportion.
While our default implementation uses \texttt{GPT-5} for demonstration construction, \method is not fundamentally restricted to an external oracle.
We evaluate alternative strategies, including self-generated and rule-based demonstrations.

\section{Experimental Setup}

\subsection{Evaluation Protocol}
Following \citet{lin-etal-2022-shot}, we use English prompts with cross-lingual 5-shot demonstrations.
For each experiment, we set 3 target languages in total---\ie French (\emph{high}), Korean (\emph{mid}), Yoruba (\emph{low})---and report performance in: (1) the target language and (2) random, unseen languages that do not appear in the demonstrations.
Evaluating unseen languages is particularly important, as real-world multilingual systems are often deployed in settings where the test language rarely appears in the training or demonstration phase.

We compare \method against five conventional X-ICL baselines following \citet{asai-etal-2024-buffet}:
\begin{itemize}[leftmargin=*, noitemsep]
    \item \textbf{Monolingual} prompting in either English or a target language~\citep{lin-etal-2022-shot, tanwar-etal-2023-multilingual};
    \item \textbf{Parallel} demonstrations in both the target language and English~\citep{kim2024crosslingual};
    \item \textbf{Translation}-based prompting from the target language into either English~\citep{huang-etal-2023-languages} or a random pivot language~\citep{chai2025xcot}.
\end{itemize}

\subsection{Evaluation Models}
We use four state-of-the-art multilingual LLMs (\ie two open and two proprietary models): \texttt{Qwen3-32B}~\citep{yang2025qwen3}, \texttt{deepseek-chat-v3.1}~\citep{deepseekai2025deepseekv3},  \texttt{grok-4-fast}~\citep{xai2025grok4fast}, and \texttt{Gemini 2.5 Flash}~\citep{comanici2025gemini}.

\subsection{Datasets and Metrics}
We use Global MMLU~\citep{singh-etal-2025-global} as a primary benchmark, covering 10 languages and diverse subject domains.
In particular, the languages include English, three target languages from three distinct language families, and six unseen languages across six different language families spanning resource levels---Chinese, Spanish (\emph{high}); Indonesian, Turkish (\emph{mid}); Swahili, Telugu (\emph{low}).

We further evaluate cross-lingual transfer across tasks and knowledge domains using five additional datasets, spanning translation, reasoning, and knowledge tasks.
We fix Spanish as the target and select one mid- to high-resource unseen language per dataset (shown in parentheses): 
\begin{itemize}[leftmargin=*, noitemsep]
    \item FLORES+~\citep{costa2024scaling} for machine translation to English (Japanese);
    \item MedExpQA~\citep{alonso2024medexpqa} for domain-specific (\ie medical) reasoning (French); 
    \item PolyMath~\citep{wang2025polymath} for mathematical reasoning (Chinese); 
    \item BLEnD~\citep{myung2024blend} for cultural knowledge (Korean); 
    \item MBBQ~\citep{neplenbroek2024mbbq} for social bias (Dutch).
\end{itemize}

We use accuracy, exact match (EM), and COMET~\citep{rei-etal-2022-comet} as evaluation metrics for multiple-choice questions (Global MMLU and MBBQ), short-answer questions (MedExpQA, PolyMath, and BLEnD), and translation (FLORES+), respectively.

\subsection{Statistical Significance Testing}
Due to limited computational resources, we conduct the following experiments as a single run with a fixed seed.
Instead, we apply bootstrap resampling with 2,000 iterations over the evaluation set to rigorously evaluate whether \method significantly outperforms the baseline systems.
In each iteration, we compute the performance difference between \method and a given baseline and derive a 95\% percentile bootstrap confidence interval (CI) from the resulting distribution.

\section{Results and Analyses}

\subsection{Main Results}
\label{sec:main_result}
Table~\ref{tab:main_result} reports the macro-averaged results of \method on Global MMLU, where the full results are stated in Appendix~\ref{sec:full_results}.
Table~\ref{tab:main_result_others} shows additional experimental results controlling the combination of demonstration and instruction settings.
\method outperforms existing X-ICL baselines in both target and unseen languages.
Specifically, \method produces 6.0pp higher performance than monolingual demonstrations in the target language, while the performance gap in English remains within 0.2pp.
In addition, \method shows substantial cross-lingual transfer in unseen, low-resource languages (+4.8pp), highlighting its practical value in multilingual scenarios.
We observe that monolingual demonstrations in the target language yield higher performance, whereas the benefit diminishes in unseen languages.
This contrast highlights a key limitation of monolingual prompting: it overfits the demonstrated language rather than fostering cross-lingual alignment. 
In contrast, \method explicitly facilitates this alignment through gradual code-switching, resulting in stronger transfer to unseen settings.
We do not observe significant differences in the \emph{out-of-format} ratio, which remains below 0.05\% for all X-ICL settings.
We further demonstrate the effectiveness of \method across model sizes in Appendix~\ref{sec:appendix_ablation} (Figure~\ref{fig:scalability}).
We also control for prompt budget and structural differences among X-ICL baselines in Appendix~\ref{sec:translation_paraphrasing}, all of which still underperform \method.

\begingroup
\setlength{\tabcolsep}{4pt}
\begin{table*}[htb!]
\centering
\resizebox{\linewidth}{!}{
\begin{tabular}{@{}l|llll|ccccc@{}}
\toprule
\multirow{2}{*}{Method}            & \multicolumn{4}{c|}{X-ICL setting}                                                                                        & \multirow{2}{*}{En} & \multirow{2}{*}{Tgt.$^*$} & \multicolumn{3}{c}{Unseen Lang.}              \\ \cmidrule(lr){2-5} \cmidrule(l){8-10} 
                                   & \multicolumn{2}{c|}{Demonstration}                                & \multicolumn{2}{c|}{Instruction}                      &                     &                       & High$^*$      & Mid$^*$       & Low$^*$       \\ \midrule
Zero-shot learning                 & \xmark & \multicolumn{1}{l|}{}                                    & \xmark &                                              & 88.6                & 68.6                  & 86.2          & 62.1          & 39.4          \\ \midrule
\multirow{3}{*}{Few-shot learning} & \cmark & \multicolumn{1}{l|}{Monolingual (En)}                    & \xmark &                                              & \textbf{88.8}       & 70.8                  & 86.5          & 62.8          & 41.2          \\
                                   & \cmark & \multicolumn{1}{l|}{Monolingual (Tgt.)}                  & \xmark &                                              & \textbf{88.8}       & 72.0                  & 86.9          & 62.1          & 38.7          \\ \cmidrule(l){2-10} 
                                   & \cmark & \multicolumn{1}{l|}{Parallel}                            & \xmark &                                              & 88.7                & 72.7                  & 87.1          & 63.0          & 41.4          \\ \midrule
\multirow{2}{*}{Zero-shot CoT}     & \xmark & \multicolumn{1}{l|}{}                                    & \cmark & Translation (Tgt.\textrightarrow En)         & \textbf{88.8}       & {\ul 74.5}            & 87.4          & 63.7          & 42.0          \\
                                   & \xmark & \multicolumn{1}{l|}{}                                    & \cmark & Translation (Tgt.\textrightarrow Rnd.)       & 88.6                & 73.8                  & {\ul 87.5}    & {\ul 63.8}    & {\ul 42.3}    \\ \midrule
\multirow{1}{*}{\method}           & \cmark & \multicolumn{1}{l|}{Gradual CS (Tgt.\textrightarrow En)} & \cmark & Gradual Translation (Tgt.\textrightarrow En) & 88.6                & \textbf{76.8}         & \textbf{87.8} & \textbf{64.9} & \textbf{46.0} \\ \bottomrule
\end{tabular}
}
\caption{Experimental results comparing \method to existing X-ICL baselines using Qwen 3. X-ICL setting describes each method in terms of a combination of its few-shot demonstrations and instruction. \method outperforms X-ICL baselines in both target and unseen languages, yielding stable results in English. \\
Throughout this paper, \textbf{bold} and {\ul underline} denote the best and the second-best results, respectively. Tgt. and Rnd. denote a target language and a random language, respectively. Asterisk indicates statistical significance against each and every baseline.}
\label{tab:main_result}
\end{table*}
\endgroup

\begin{figure}[htb!]
\centering

\begin{minipage}[c][5cm][c]{.35\linewidth}
    \centering
    \vfill
    \includegraphics[width=\linewidth]{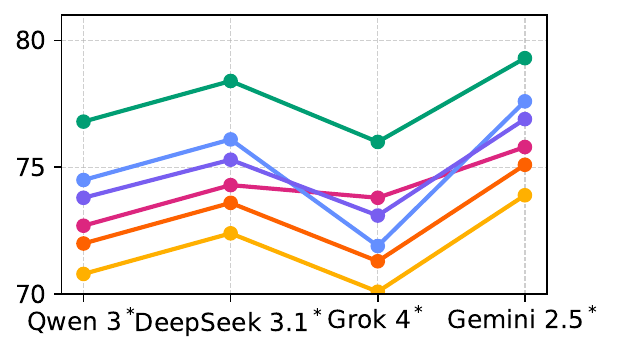}
    \vfill
    \subcaption{Models}
    \label{fig:per_model}
\end{minipage}
\hfill
\begin{minipage}[c][5cm][c]{.35\linewidth}
    \centering
    \includegraphics[width=\linewidth]{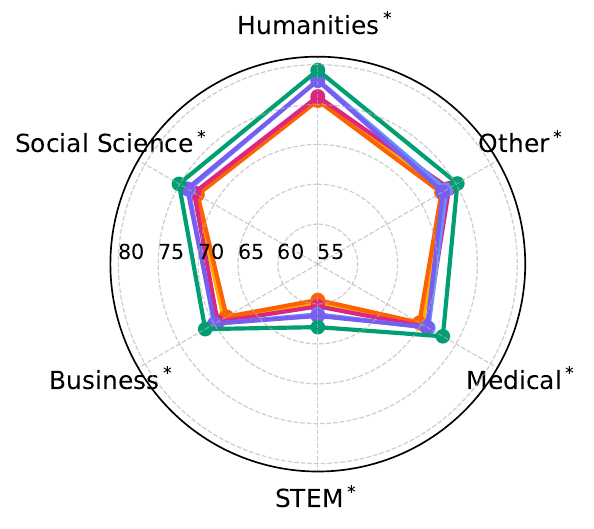}
    \vfill
    \subcaption{Subject categories}
    \label{fig:per_subject_qwen}
\end{minipage}
\hfill
\begin{minipage}[c][5cm][c]{.25\linewidth}
    \centering
    \includegraphics[width=\linewidth]{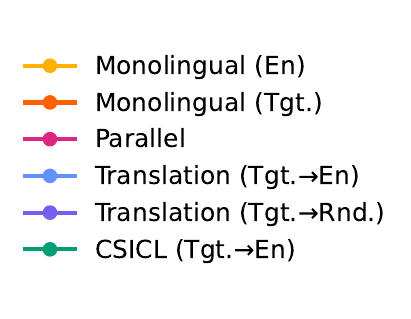}
\end{minipage}

\caption{Experimental results of \method compared to X-ICL approaches.}
\label{fig:model_and_subject}
\end{figure}

\paragraph{Models.}
Figure~\ref{fig:per_model} displays experimental results of each model with X-ICL approaches in target languages on Global MMLU.
All four models achieve similar trends across X-ICL settings, where \method significantly outperforms existing baselines.

\paragraph{Subject categories.}
Figure~\ref{fig:per_subject_qwen} shows experimental results of X-ICL approaches using Qwen 3 for six subject categories on Global MMLU.
\method significantly surpasses baselines across all categories.

\paragraph{Languages.}

\begin{figure}[htb!]
    \centering
    \subfloat{\centering\includegraphics[width=\linewidth]{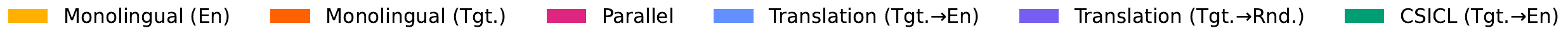}}
    
    \addtocounter{subfigure}{-1}
    \subfloat[\centering Target: French (\emph{high})]{\includegraphics[width=.33\linewidth]{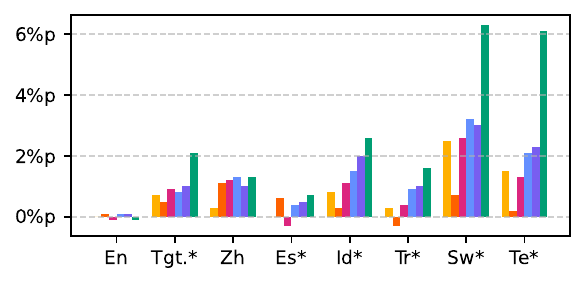}}
    \hfill
    \subfloat[\centering Target: Korean (\emph{mid})]{\includegraphics[width=.33\linewidth]{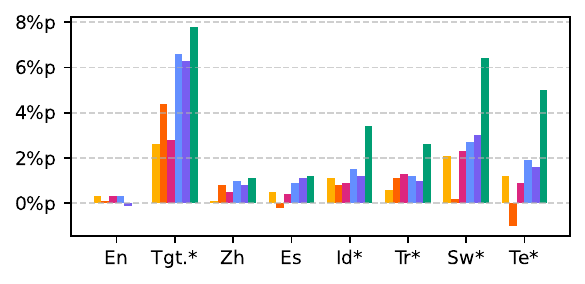}}
    \hfill
    \subfloat[\centering Target: Yoruba (\emph{low})]{\includegraphics[width=.33\linewidth]{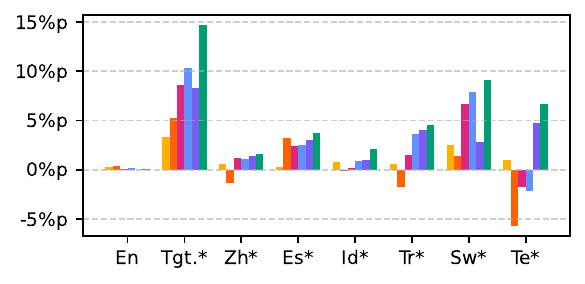}}
    \caption{Performance differences (pp) of \method and X-ICL baselines compared to zero-shot learning setting per a target language using Qwen 3.}
    \label{fig:per_language}
\end{figure}

Figure~\ref{fig:per_language} presents performance differences (pp) of each X-ICL setting compared to zero-shot learning, whose target languages are French, Korean, and Yoruba, respectively.
\method outperforms X-ICL baselines, particularly in target languages and mid- to low-resource unseen languages, while English performance remains stable.
The benefits of \method extend beyond script similarity or language family, implying that it fosters language-agnostic cross-lingual alignment rather than relying on superficial linguistic overlap.

\subsection{Task and Domain Knowledge}
\label{sec:task_ablation}

Table~\ref{tab:task_result} presents experimental results of X-ICL approaches using Qwen 3 across five additional tasks spanning translation, reasoning, and knowledge evaluation.
Full experimental results with the other three models are presented in Appendix~\ref{sec:full_results}.
In line with earlier results, \method consistently outperforms existing baselines across settings.
Following the descriptions provided in the original papers, we categorize tasks into three types: machine translation (FLORES+); reasoning-oriented tasks (MedExpQA, PolyMath); and knowledge-intensive tasks (BLEnD, MBBQ).

\begingroup
\setlength{\tabcolsep}{4pt}
\begin{table*}[htb!]
\centering
\resizebox{\linewidth}{!}{
\begin{tabular}{@{}l|cc|cccccc|cccccc@{}}
\toprule
\multirow{2}{*}{Task type}             & \multicolumn{2}{c|}{\multirow{2}{*}{Translation}} & \multicolumn{6}{c|}{Reasoning-oriented}                                                                               & \multicolumn{6}{c}{Knowledge-intensive}                                                                                  \\ \cmidrule(l){4-15} 
                                       & \multicolumn{2}{c|}{}   & \multicolumn{3}{c|}{Domain-specific}                               & \multicolumn{3}{c|}{Math}                     & \multicolumn{3}{c|}{Cultural}                                            & \multicolumn{3}{c}{Social Bias}               \\ \midrule
X-ICL setting                          & Tgt.$^*$      & Uns.$^*$      & En            & Tgt.$^*$      & \multicolumn{1}{c|}{Uns.$^*$}          & En            & Tgt.$^*$      & Uns.$^*$      & En            & Tgt.$^*$      & \multicolumn{1}{c|}{Uns.$^*$}                & En            & Tgt.$^*$      & Uns.          \\ \midrule
Monolingual (En)                       & 76.6          & 72.3          & \textbf{46.4} & 38.9          & \multicolumn{1}{c|}{32.1}          & {\ul 51.2}    & 49.3          & 47.4          & {\ul 83.7}    & 77.8          & \multicolumn{1}{c|}{73.2}                & {\ul 92.0}    & 88.9          & 86.3          \\
Monolingual (Tgt.)                     & 75.3          & 71.6          & 45.8          & 38.5          & \multicolumn{1}{c|}{30.3}          & 50.5          & 49.7          & 46.1          & 83.1          & 78.9          & \multicolumn{1}{c|}{71.5}                & 91.6          & 89.3          & 85.8          \\ \midrule
Parallel                               & {\ul 78.1}    & 73.0          & {\ul 46.3}    & {\ul 40.2}    & \multicolumn{1}{c|}{31.9}          & \textbf{51.4} & 49.7          & 47.7          & \textbf{83.8} & {\ul 79.2}    & \multicolumn{1}{c|}{72.0}                & 91.5          & 89.5          & 86.0          \\ \midrule
Translation (Tgt.\textrightarrow En)   & 74.8          & 72.1          & 46.1          & 39.6          & \multicolumn{1}{c|}{32.3}          & 50.8          & 50.8          & {\ul 48.2}    & 83.3          & 78.5          & \multicolumn{1}{c|}{71.7}                & \textbf{92.1} & {\ul 90.1}    & 86.4          \\
Translation (Tgt.\textrightarrow Rnd.) & 73.2          & {\ul 73.7}    & {\ul 46.3}    & 40.1          & \multicolumn{1}{c|}{{\ul 32.7}}    & 50.3          & {\ul 51.0}    & 47.9          & 83.6          & 78.1          & \multicolumn{1}{c|}{{\ul 72.4}}          & 91.9          & 89.8          & \textbf{87.3} \\ \midrule
\method (Tgt.\textrightarrow En)       & \textbf{83.4} & \textbf{75.3} & 46.2          & \textbf{44.4} & \multicolumn{1}{c|}{\textbf{35.2}} & 50.9          & \textbf{54.5} & \textbf{51.8} & 83.5          & \textbf{81.3} & \multicolumn{1}{c|}{{\ul \textbf{74.5}}} & 91.8          & \textbf{90.5} & {\ul 87.2}    \\ \bottomrule
\end{tabular}
}
\caption{Experimental results on 5 diverse tasks and domain knowledge using Qwen 3. \method achieves larger improvements in translation and reasoning tasks.}
\label{tab:task_result}
\end{table*}
\endgroup

\paragraph{Translation.}
\method achieves the largest gains in the translation task---gains of +6.8pp and +3.0pp COMET in the target and the unseen languages, respectively, over monolingual demonstrations. 
This strong effect is expected, as progressive transition directly scaffolds the latent translation process, aligning the input with English-centric representations.
It is noteworthy that \method is effective even in a purely generative setting.

\paragraph{Reasoning.}
Beyond translation, \method also improves reasoning-oriented tasks, with average gains of +5.4pp and +3.8pp of EM over monolingual baselines. 
We suppose that gradual transitions mitigate representational interference: instead of reasoning entirely in a non-English space, the demonstrations nudge the model to ``\emph{think in English,}'' where reasoning capabilities are strongest. 
This highlights the potential of \method as a lightweight test-time alignment mechanism, complementary to scaling methods such as self-consistency.

\paragraph{Knowledge.}
For cultural knowledge and social bias, the improvements are modest but still consistent (+2.6pp and +1.1pp in the target and the unseen languages, respectively, on average).
This indicates that \method not only facilitates translation-like tasks but also improves general factual consistency across languages.

\subsection{What Makes \method Effective?}

\begin{figure*}[htb!]
    \centering
    \subfloat{\centering\includegraphics[width=0.45\linewidth]{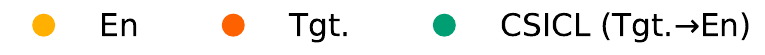}}
    
    \addtocounter{subfigure}{-1}
    
    \subfloat[\centering Layer 1]{\includegraphics[width=0.19\linewidth]{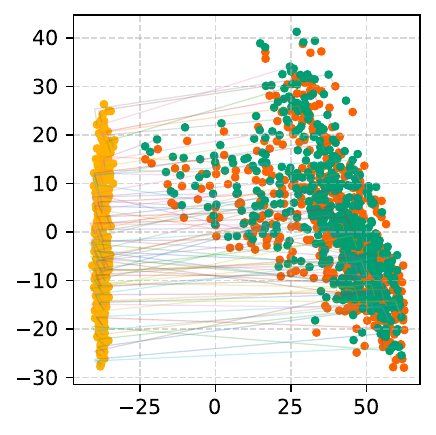}}
    \hfill
    \subfloat[\centering Layer 8]{\includegraphics[width=0.19\linewidth]{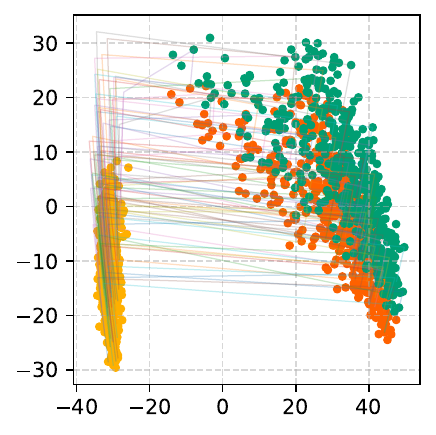}}
    \hfill
    \subfloat[\centering Layer 16]{\includegraphics[width=0.19\linewidth]{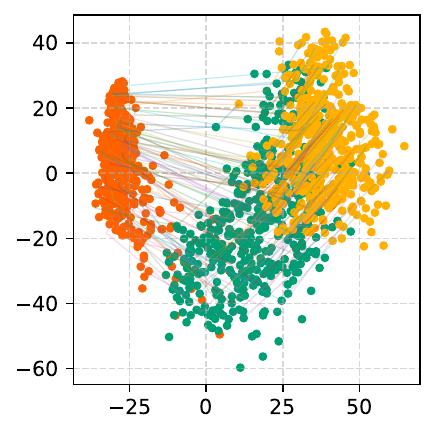}}
    \hfill
    \subfloat[\centering Layer 32]{\includegraphics[width=0.19\linewidth]{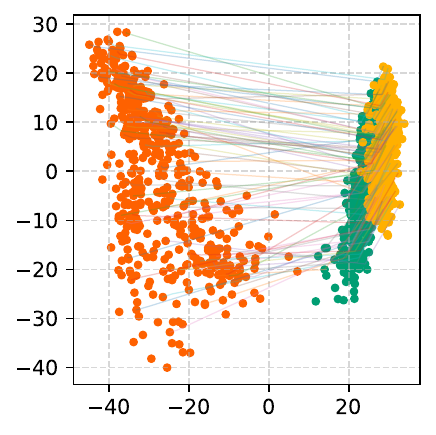}}
    \hfill
    \subfloat[\centering Layer 64]{\includegraphics[width=0.19\linewidth]{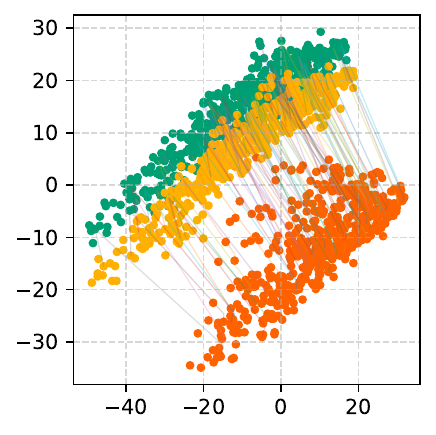}}
    \caption{PCA visualizations of Qwen 3 with \method on Global MMLU, compared to English results and Korean results as anchors. }
    \label{fig:pca}
\end{figure*}

\paragraph{Layer-wise visualization of \method.}
\label{sec:pca}
We analyze whether \method helps an LLM resolve cross-lingual misalignment by explicitly facilitating its latent transition to English representations.
We evaluate three input conditions: Korean (Ko), English equivalent (En), and \method applied to the same Korean set (\method).
For each condition, we run forward passes and extract hidden representations from every layer, focusing on the test query span without few-shot contexts via span-based pooling to obtain a single vector per layer per example.
We then visualize how these representations evolve across depth by projecting layer-wise vectors with PCA, plotting matched triples (\ie En, Ko, \method) in a shared space.
Figure~\ref{fig:pca} visualizes the representations of \texttt{Qwen3-32B}, compared to English and Korean results as anchors.
We reveal that \method systematically shifts the non-English inputs toward the English anchor.

We further examine which design factors contribute to the gains of \method and report the experimental results on Global MMLU using Qwen 3 in Table~\ref{tab:ablation_result}.

\paragraph{Demonstrations vs. instruction.}
We observe that both gradual code-switching demonstrations and translation instruction contribute to cross-lingual transfer, by yielding 3.4pp and 3.8\% higher accuracy compared to the monolingual baseline.
In particular, gradual code-switching consistently achieves higher performances than code-switching (+0.5pp), parallel (+1.5pp), and monolingual demonstrations (+3.4pp).
Zero-shot and gradual translation instructions yield minimal differences within 0.1pp in the target language, as the model fails to understand the concept of gradual translation and to follow it without explicit demonstrations.
Specifically, zero-shot gradual translation instruction often collapses into abrupt translation, where the model outputs two sentences entirely in Korean, followed by three equivalent sentences entirely in English.

\paragraph{Directionality.}
Furthermore, transitioning from a target language to English outperforms the reverse direction, supporting our hypothesis that \method alleviates the translation barrier by scaffolding the latent translation process of LLMs. 
This asymmetry indicates that LLMs benefit more when demonstrations converge to English representations, aligning with their latent space, rather than diverging away from it.

\begingroup
\setlength{\tabcolsep}{4pt}
\begin{table*}[h]
\centering
\resizebox{\linewidth}{!}{
\begin{tabular}{@{}l|llll|ccccc@{}}
\toprule
\multirow{2}{*}{Method}            & \multicolumn{4}{c|}{X-ICL setting}                                                                                        & \multirow{2}{*}{En} & \multirow{2}{*}{Tgt.$^*$} & \multicolumn{3}{c}{Unseen Lang.}              \\ \cmidrule(lr){2-5} \cmidrule(l){8-10} 
                                   & \multicolumn{2}{c|}{Demonstration}                                & \multicolumn{2}{c|}{Instruction}                      &                     &                       & High$^*$      & Mid$^*$       & Low$^*$       \\ \midrule
\multirow{4}{*}{Few-shot learning} & \cmark & \multicolumn{1}{l|}{CS (En+Tgt.)}                        & \xmark &                                              & 88.6                & 73.6                  & 87.1          & 62.9          & 43.3          \\
                                   & \cmark & \multicolumn{1}{l|}{CS (Tgt.+En)}                        & \xmark &                                              & 88.7                & 73.7                  & 87.0          & 62.8          & 43.1          \\ \cmidrule(l){2-10} 
                                   & \cmark & \multicolumn{1}{l|}{Gradual CS (En\textrightarrow Tgt.)} & \xmark &                                              & 88.7                & 74.0                  & 87.6          & 63.8          & 44.5          \\
                                   & \cmark & \multicolumn{1}{l|}{Gradual CS (Tgt.\textrightarrow En)} & \xmark &                                              & 88.6                & 74.2                  & {\ul 87.7}    & {\ul 64.3}    & {\ul 45.7}    \\ \midrule
\multirow{1}{*}{Zero-shot CoT}     & \xmark & \multicolumn{1}{l|}{}                                    & \cmark & Gradual Translation (Tgt.\textrightarrow En) & 88.7                & 74.6                  & 87.2          & 63.8          & 42.0          \\ \midrule
\multirow{2}{*}{\method}           & \cmark & \multicolumn{1}{l|}{Gradual CS (En\textrightarrow Tgt.)} & \cmark & Gradual Translation (En\textrightarrow Tgt.) & 88.7                & {\ul 75.0}            & 87.6          & 64.0          & 45.2          \\
                                   & \cmark & \multicolumn{1}{l|}{Gradual CS (Tgt.\textrightarrow En)} & \cmark & Gradual Translation (Tgt.\textrightarrow En) & 88.6                & \textbf{76.8}         & \textbf{87.8} & \textbf{64.9} & \textbf{46.0} \\ \bottomrule
\end{tabular}
}
\caption{Ablation results to examine the effectiveness of each design factor in \method. Gradual transition demonstrations, instruction, and direction contribute to cross-lingual transfer.}
\label{tab:ablation_result}
\end{table*}
\endgroup

\paragraph{Transition granularity.}

Table~\ref{tab:granularity} shows experimental results of \method on Global MMLU using Qwen 3, controlling the number of gradual transition steps and the number of few-shot demonstrations.
We also compare linear vs. non-linear schedules of gradual code-switching.
We observe that the more steps in gradual transition and the larger number of demonstrations result in better target and unseen language performance.
However, the performance gap between 5-step and 7-step transitions is minimal, and even mixed.
This suggests that excessively fine-grained switching may exceed the model's controllable generation capability, while gradual alignment is beneficial.
Similarly, performance increases up to 5 shots and then saturates, indicating that gradual transition behavior stabilizes once sufficiently demonstrated.
Additionally, linear transition consistently outperforms non-linear variants, indicating that smooth interpolation is more effective than abrupt jumps.

\begin{table}[htb!]
    \centering
    \begin{subtable}[t]{0.52\linewidth}
        \resizebox{\linewidth}{!}{
\begin{tabular}[t]{@{}lccccc@{}}
\toprule
\multirow{2}{*}{\method (5 shots)} & \multirow{2}{*}{En} & \multirow{2}{*}{Tgt.} & \multicolumn{3}{c}{Unseen Lang.}              \\ \cmidrule(l){4-6}
                                   &                     &                       & High          & Mid           & Low           \\ \midrule
0-25-100 (3 steps)                 & {\ul 88.6}          & 74.5                  & 87.0          & 63.9          & 42.4          \\
0-75-100 (3 steps)                 & {\ul 88.6}          & 74.3                  & 87.2          & 63.8          & 42.3          \\
0-50-100 (3 steps)                 & \textbf{88.7}       & 75.1                  & {\ul 87.7}    & 64.0          & 44.4          \\
0-25-75-100 (4 steps)              & {\ul 88.6}          & 76.0                  & {\ul 87.7}    & 64.5          & 45.3          \\
0-25-50-75-100 (5 steps)           & {\ul 88.6}          & {\ul 76.8}            & \textbf{87.8} & {\ul 64.9}    & \textbf{46.0} \\
0-16-33-50-66-83-100 (7 steps)     & 88.5                & \textbf{76.9}         & 87.5          & \textbf{65.1} & {\ul 45.8}    \\ \bottomrule
\end{tabular}
}
\caption{\# of steps}
\label{tab:granularity_steps}
    \end{subtable}
    \hfill
    \begin{subtable}[t]{0.47\linewidth}
        \resizebox{\linewidth}{!}{
\begin{tabular}[t]{@{}lccccc@{}}
\toprule
\multirow{2}{*}{\method (5 steps)} & \multirow{2}{*}{En} & \multirow{2}{*}{Tgt.} & \multicolumn{3}{c}{Unseen Lang.}              \\ \cmidrule(l){4-6} 
                                   &                     &                       & High          & Mid           & Low           \\ \midrule
0-25-50-75-100 (0 shot)            & \textbf{88.7}       & 74.6                  & 87.2          & 63.8          & 42.0          \\
0-25-50-75-100 (1 shot)            & \textbf{88.7}       & 75.3                  & 87.3          & 64.2          & 44.6          \\
0-25-50-75-100 (3 shots)           & {\ul 88.8}          & 76.2                  & 87.6          & 64.7          & 45.3          \\
0-25-50-75-100 (5 shots)           & {\ul 88.6}          & {\ul 76.8}            & \textbf{87.8} & {\ul 64.9}    & {\ul 46.0}    \\
0-25-50-75-100 (7 shots)           & {\ul 88.6}          & \textbf{77.0}         & {\ul 87.7}    & \textbf{65.0} & \textbf{46.1} \\ \bottomrule
\end{tabular}
}
\caption{\# of demonstrations}
\label{tab:granularity_shots}
    \end{subtable}
    \caption{Experimental results of \method using Qwen 3, controlling the number of transition steps and the number of few-shot demonstrations.}
    \label{tab:granularity}
\end{table}

\subsection{Is \method Practically Useful?}
\label{sec:practical}

\paragraph{Cost--performance trade-off.}

To quantify the cost--performance trade-off, we compute the average number of input/output tokens for X-ICL approaches using the \texttt{Qwen3-32B} tokenizer on Global MMLU.
Table~\ref{tab:cost_benefit} summarizes cost--performance trade-off.
As expected, the full \method configuration with five few-shot demonstrations and a five-step transition uses more tokens than simpler X-ICL baselines, while yielding the highest gains.
However, its average token budget of 1,446.7 remains well within practical context limits for Qwen 3, whose native context window is 32,768 tokens and can be extended to 131,072 with YaRN.
While longer prompts directly affect latency and costs in API-based deployments, the overhead is comparable to that of other few-shot approaches rather than an order-of-magnitude increase.
Importantly, our ablation with fewer-shot variants shows that even 0-shot and 1-shot versions of \method already outperform strong X-ICL baselines while using fewer tokens than existing X-ICL alternatives.
These results suggest that \method offers a favorable and tunable efficiency--accuracy trade-off: 0--1 shot is attractive for budget-constrained settings, whereas 3+ shots are preferable when maximizing accuracy is the priority.

\begin{table}[htb!]
\centering
\resizebox{0.8\linewidth}{!}{
\begin{tabular}{@{}lccc@{}}
\toprule
X-ICL setting         & Avg. input tokens & Avg. output tokens & Tgt. performance \\ \midrule
Zero-shot learning    & 86.9              & 4.8                & 68.6             \\
Parallel              & 651.7             & 131.6              & 72.7             \\
Translation (Tgt.→En) & 438.8             & 133.2              & 74.5             \\ \midrule
\method (0 shot)      & 231.5             & 154.7              & 74.6             \\
\method (1 shot)      & 387.6             & 184                & 75.3             \\
\method (3 shots)     & 868.4             & 181.8              & 76.2             \\
\method (5 shots)     & 1,265.1           & 181.6              & 76.8             \\ \bottomrule
\end{tabular}
}
\caption{Cost--performance trade-off of X-ICL approaches considering both required tokens and target language performance.}
\label{tab:cost_benefit}
\end{table}

\paragraph{Demonstration generation.}

\begin{wraptable}{R}{0.5\linewidth}
\vspace{-4mm}
\centering
\resizebox{\linewidth}{!}{
\begin{tabular}{@{}lccccc@{}}
\toprule
\multirow{2}{*}{Demonstrations} & \multirow{2}{*}{En} & \multirow{2}{*}{Tgt.} & \multicolumn{3}{c}{Unseen Lang.} \\ \cmidrule(l){4-6} 
                                      &                     &                       & High      & Mid       & Low      \\ \midrule
\texttt{GPT-5}                        & 88.6                & 76.8                  & 87.8      & 64.9      & 46.0     \\ \midrule
Same model             & 88.7                & 76.7                  & 87.8      & 64.6      & 45.8     \\
SLM                   & 88.6                & 75.5                  & 87.5      & 64.2      & 44.3     \\
Rule-based                            & 88.6                & 75.2                  & 87.5      & 64.4      & 44.0     \\ \bottomrule
\end{tabular}
}
\caption{Experimental results of \method using Qwen 3 with few-shot demonstrations generated by three alternative strategies: (1) the same model of the X-ICL inference setup, (2) a small language model (SLM), and (3) a rule-based method without any external model.}
\label{tab:alternatives}
\end{wraptable}

We examine whether CSICL depends on a frontier model for few-shot demonstration generation.
Here, we set up three alternative generation strategies: (1) using the same model (\texttt{Qwen3-32B}) of the X-ICL inference setup, (2) a small language model (\texttt{gemma-3-1b}) to lower the cost for few-shot generation, and (3) a rule-based method without relying on any external model.
Table~\ref{tab:alternatives} reports the experimental results of \method with alternative strategies for few-shot demonstration generation.
Self-generated demonstrations achieve nearly identical performance to \texttt{GPT-5}.
This implies that \method enables a truly end-to-end multilingual system regardless of whether external models are used.
While small model and rule-based approaches yield slightly lower performance, they still outperform other X-ICL baselines. This demonstrates and supports the feasibility and robustness of \method.
Importantly, \method requires only five demonstrations to improve 14.7pp of target language performance.
These handful of demonstrations can be produced automatically, as we have shown above, or even manually curated with minimal effort. 

\section{Related Work}
\subsection{Cross-lingual Transfer in LLMs}

Recent studies suggest that LLMs tend to rely on English as a pivot for cross-lingual processing.
\citet{zhao2024how} showed that LLMs exhibit language-specific neurons to process multilingual inputs, and
\citet{schut2025multilingual} demonstrated that LLMs internally translate multilingual inputs into English representations and think in English.
This reliance on internal translation introduces a critical vulnerability: \emph{translation barrier}, where a failure in the initial translation stage is propagated to the quality of the final outputs~\citep{bafna2025translation}.
While external machine translation can slightly mitigate this issue~\citep{etxaniz-etal-2024-multilingual, intrator-etal-2024-breaking}, it undermines the goal of using LLMs as seamless, end-to-end multilingual systems.
This motivates the exploration of mechanisms that can directly activate cross-lingual capabilities of LLMs without relying on latent English translation.

To bridge this gap, cross-lingual transfer has been widely studied~\citep{pallucchini2025lost}.
For example, \citet{tanwar-etal-2023-multilingual, huang-etal-2023-languages, li-etal-2024-improving-context, lin-etal-2025-xampler} investigated X-ICL approaches to better align cross-lingual representations.
However, these approaches are primarily based on an abrupt single-step pivot, requiring a more effective cross-lingual signal that can mitigate this misalignment during inference.

\subsection{Code-switching for Cross-lingual Transfer}
Code-switching, which has been studied over decades by the natural language processing community~\citep{dogruoz-etal-2021-survey, winata-etal-2023-decades}, is a promising candidate for such a direct signal.
Code-switching has shown its efficacy for cross-lingual transfer in various training stages, including pre-training~\citep{wang-etal-2025-investigating-scaling}, continual pre-training~\citep{yoo-etal-2025-code-switching}, supervised fine-tuning (SFT)~\citep{lee-etal-2024-commit, chai2025xcot, son2026pushing}, and selective constrained decoding~\citep{li2025impact}.
In particular, \citet{chai2025xcot} demonstrated that (1) SFT using inter-sentential code-switching data and (2) chain-of-thought prompting, which translates a non-English query into a random language and answers in the target language, enhances cross-lingual transfer.
\citet{li2025impact} introduced probe-guided decoding with an additional classifier to enhance bilingual LLM reasoning with a case study of English-Chinese code-switching.
\citet{son2026pushing} also demonstrated that SFT using code-switching instruction data enhances LLM reasoning.
\citet{kang-etal-2026-multilingual} suggested to selectively incorporate English translations into the latent reasoning of LLMs using additional classifiers.

However, these approaches require substantial additional training resources, while exploration remains limited at the inference level, often focusing only on a specific task and domain knowledge.
For machine reading comprehension, \citet{kim2024crosslingual} provided a passage in English and a QA pair in a target language, which is a form of intra-sentential code-switching.
\citet{kim2025codeswitched} showcased a case study where Korean-English inter-sentential code-switching can unlock Korean cultural knowledge that remains inaccessible in English queries.
\citet{wang-etal-2025-language-mixing} investigated patterns, impact, and internal causes of code-switching in reasoning language models.
Our work, in contrast, moves beyond utilizing code-switching merely as a static context.
We introduce an inference-time representational alignment mechanism that systematically controls the reasoning trajectory.
By gradually transitioning from the target language to English, \method explicitly scaffolds the latent space of LLMs, improving cross-lingual transfer during inference.

\section{Conclusion}
We address cross-lingual misalignment in multilingual LLMs---their over-reliance on English-centric latent representations that often leads to degraded performance in non-English.
To mitigate this issue, we introduce \method, an inference-time method that improves cross-lingual transfer by gradually transitioning from a target language to English through structured code-switching.
Across 4 multilingual LLMs, 6 datasets, and 10 languages,
\method consistently outperforms standard X-ICL baselines, yielding average gains of 6.0pp and 4.8pp in both target and unseen languages, respectively.
We demonstrate its pronounced improvements in low-resource settings, with gains of 14.7pp in target and 5.3pp in unseen languages.
Beyond empirical improvements, this paper positions gradual code-switching as a new lens for cross-lingual alignment during LLM inference.

\section*{Acknowledgments}
We thank Dongkwan Kim and Guijin Son for their valuable advice.

This work was supported by Institute of Information \& communications Technology Planning \& Evaluation(IITP) grant funded by the Korea government(MSIT) (No. RS-2024-00509258 and No. RS-2024-00469482, Global AI Frontier Lab).
This research was conducted as part of the Sovereign AI Foundation Model Project(GPU Track), organized by the Ministry of Science and ICT(MSIT) and supported by the National IT Industry Promotion Agency(NIPA), S.Korea. (PJT-26-010017).
This research was supported by the MSIT(Ministry of Science, ICT), Korea, under the Top-Tier AI Global HRD invitation program (RS-2025-25461932) supervised by the IITP(Institute for Information \& Communications Technology Planning \& Evaluation).

We use ChatGPT, Gemini, and Copilot for writing and coding assistance.
We also use LLMs to generate few-shot demonstrations in the experiments (\eg gradual code-switching, translation, and paraphrasing).

\section*{Ethics Statement}
All datasets used in this study are publicly available and employed solely for research purposes, consistent with their intended licenses. 
We prevent data leakage by excluding the samples used as few-shot demonstrations from the corresponding test set.
\method aims to improve the multilingual performance of LLMs, with the intended use for both research and practical applications, which is not used for malicious purposes.

\section*{Reproducibility Statement}
For open-source LLM inference, we use 4 H200 GPUs with 564 GB memory and 4 RTX 8000 GPUs with 188 GB memory.
For closed-source LLMs, we access them via OpenRouter\thinspace\footnote{\url{https://openrouter.ai/}}.
All experiments and sample selections are conducted with a random seed of 42.
All COMET scores in this paper are COMET-22 (\texttt{unbabel/wmt22-comet-da})~\citep{rei-etal-2022-comet}.
For reliable reproduction, we use greedy decoding with a temperature of 0.0, if possible.
All system prompts used in the experiments are provided in Section~\ref{sec:system_prompts}.

All evaluation datasets used in this paper are publicly available.
Global MMLU~\citep{singh-etal-2025-global} is released under the \texttt{Apache-2.0} license. 
BLEnD~\citep{myung2024blend}, FLORES+~\citep{costa2024scaling}, and MedExpQA~\citep{alonso2024medexpqa} are distributed under the \texttt{CC-BY-SA-4.0} and \texttt{CC-BY-4.0} licenses, respectively.
The licenses of MBBQ~\citep{neplenbroek2024mbbq} and PolyMath~\citep{wang2025polymath} are not explicitly specified.
The samples used as few-shot demonstrations are excluded from the corresponding test set to avoid data leakage.

We use a balanced set of 36,000 instances from Global MMLU by randomly sampling 600 questions per subject category and language.
For MBBQ, we randomly select ten samples (five ambiguous contexts and five unambiguous contexts) per template in the MBBQ dataset, totaling 980 samples, due to limited computational resources for processing the full 10k+ instances.
We employ the entire set for the other datasets. 

In Appendix~\ref{sec:practical}, we follow the experimental setup of \citet{yoo-etal-2025-code-switching} for the rule-based method.
Specifically, we employ a part-of-speech (POS) tagger and randomly replace a specified proportion of Korean nouns in Korean sentences with English words using a bilingual dictionary.
In Appendix~\ref{sec:translation_paraphrasing}, we follow the implementations of \citet{huang-etal-2023-languages} for the \emph{translate-then-reason} baseline.

\bibliography{custom}
\bibliographystyle{colm2026_conference}

\clearpage
\appendix
\section*{Appendix}

\section{Limitations}
Due to computational constraints, most experiments are reported with a single seed; the bootstrap intervals therefore capture uncertainty over evaluation instances rather than full run-to-run variability.

For evaluation, we largely employ automated metrics (\ie accuracy, exact match, and COMET), which could be considered a limitation.
We adopt automated metrics for consistency and scalability across multilingual settings, which is central to our study.
Further, a small-scale human evaluation in Appendix~\ref{sec:human_validation} shows strong agreement with the COMET-based results, indicating that our main findings are robust.

We conduct experiments on 10 languages spanning different resource levels.
While this set cannot cover all language families, it offers a broad and representative sample that supports the generality of \method.

In our task- and domain-specific evaluation (\S\ref{sec:task_ablation}), we fix Spanish as the target language and vary the unseen language across datasets due to the limited availability of parallel data in specialized tasks and domains.
While this design enables broad coverage of domains, part of the performance differences observed across tasks may stem from the choice of unseen languages rather than from the tasks themselves.
It reflects the practical challenges of multilingual evaluation and still provides a diverse and informative basis for analysis.

\section{System Prompts}
\label{sec:system_prompts}

To ground the following system prompts, we carefully curate the real-world code-switching examples from \citet{finer2014movement}.

\subsection{Few-shot Demonstrations Generation}

\begin{tcolorbox}[breakable, enhanced, top=1pt, left=1pt, right=1pt, bottom=1pt, colback=white, title=Prompt for generating English-Korean code-switching few-shot demonstrations]
\small
You are a bilingual rewriting assistant.\\
\\
{[TASK]}\\
  • Input  : an English sentence (E) and its Korean translation (K)\\
  • Output : the code-switching version of the parallel sentences following Matrix Language Frame (MLF) model\\
            - Replace about 50\% of words/phrases in E with their Korean equivalents taken from K\\
            - Keep the original English word order and follow English syntax (S-V-O)\\
            - DO NOT add explanations, examples, tags, prefix or extra sentences\\
            - If there is no suitable Korean equivalent, keep the English word\\

{[EXAMPLE]}\\
<English> I ate dinner quickly.\\
<Korean> 나는 저녁을 빨리 먹었다.\\
<Code-Switching> I ate 저녁 빨리.\\
\\
<English> Hana, put the toys in the basket quickly and go home.\\
<Korean> 하나야, 바구니에 장난감을 빨리 넣고 집에 가자.\\
<Code-Switching> Hana, put 장난감 in the basket quickly and 집에 가자.\\
\\
<English> Dad was about to throw away my tooth.\\
<Korean> 아빠가 내 이빨을 빼려고 했어.\\
<Code-Switching> 아빠 was about to 뺄래 my 이빨.\\
\\
<English> I have to wash my hand.\\
<Korean> 나는 손을 씻어야 해.\\
<Code-Switching> I have to 닦아 my hand.\\
\\
<English> Tom thinks Bill likes himself.\\
<Korean> 톰은 빌이 자기 자신을 좋아한다고 생각한다.\\
<Code-Switching> Tom thinks that Bill이 자기를 좋아한다.\\
\\
<English> Tom thinks Bill likes himself.\\
<Korean> 톰은 빌이 자기 자신을 좋아한다고 생각한다.\\
<Code-Switching> Tom이 생각하기를 Bill likes himself.\\
\\
{[BEGIN TASK]}
\end{tcolorbox}

\begin{tcolorbox}[breakable, enhanced, top=1pt, left=1pt, right=1pt, bottom=1pt, colback=white, title=Prompt for generating Korean-English code-switching few-shot demonstrations]
\small
You are a bilingual rewriting assistant.\\
\\
{[TASK]}\\
  • Input  : an English sentence (E) and its Korean translation (K)\\
  • Output : the code-switching version of the parallel sentences following Matrix Language Frame (MLF) model\\
            - Replace about 50\% of words/phrases in K with their English equivalents taken from E\\
            - Keep the original Korean word order and follow Korean syntax (S-O-V)\\
            - DO NOT add explanations, examples, tags, prefix or extra sentences\\
            - If there is no suitable English equivalent, keep the Korean word\\
\\
{[EXAMPLE]}\\
<English> Meena, put all the toys in the basket quickly, and go home.\\
<Korean> 미나야, 바구니에 장난감을 다 넣고 빨리 집에 가자.\\
<Code-Switching> Meena, basket 안에다 all the toys를 빨리 put하고 집에 가자.\\
\\
<English> Last time, by mistake, they did not renew my driver's license at the DMV.\\
<Korean> 지난 번에 도로교통공단이 내 운전면허증을 실수로 갱신하지 않았어요.\\
<Code-Switching> 지난번에 motor vehicle department에서 내 driver's license를 mistake로 갱신하지 않았어요.\\
\\
<English> They often do a thing like that.\\
<Korean> 그들이 일을 종종 그렇게 해요.\\
<Code-Switching> 그들이 일을 often 그렇게 해요.\\
\\
<English> Tom thinks Bill likes himself.\\
<Korean> 톰은 빌이 자기 자신을 좋아한다고 생각한다.\\
<Code-Switching> Tom은 Bill이 himself를 좋아한다고 생각한다.\\
\\
<English> John wonders what Mary bought yesterday.\\
<Korean> 존은 메리가 어제 무엇을 샀는지 궁금해한다.\\
<Code-Switching> John은 Mary가 yesterday 무엇을 샀는지 궁금해한다.\\
\\
{[BEGIN TASK]}
\end{tcolorbox}

\begin{tcolorbox}[breakable, enhanced, top=1pt, left=1pt, right=1pt, bottom=1pt, colback=white, title=Prompt for generating gradual code-switching (En\textrightarrow Ko) few-shot demonstrations]
\small
You are a bilingual rewriting assistant.\\
Your task is to generate five versions of a sentence that gradually transition from English to Korean.\\
\\
{[INPUT]}\\
  • One English sentence (E)\\
  • Its Korean translation (K)\\
  • A code-switching version of the sentence (C), where about 50\% of English words are replaced by Korean equivalents\\
\\
{[OUTPUT]}\\
Generate a sequence of five sentences showing a smooth progression from English to Korean:\\
  1. English only (100\% English, source syntax S-V-O)\\
  2. 75\% English + 25\% Korean (matrix language: English, embedded language: Korean)\\
  3. 50\% English + 50\% Korean (matrix language: English, embedded language: Korean)\\
  4. 25\% English + 75\% Korean (matrix language: English, embedded language: Korean)\\
  5. Korean only (100\% Korean, target syntax S-O-V)\\
\\
{[RULES]}\\
Following the Matrix Language Frame (MLF) model,\\
  • Preserve English word order (S-V-O) and syntax until version 5 (full Korean).\\
  • Use Korean equivalents from K when inserting Korean into English sentences.\\
  • Keep the code-switching natural and consistent, not random.\\
  • Do not add explanations, notes, or extra text — output only the five sentences in order.\\
\\
{[EXAMPLES]}\\
\\
EXAMPLE 1\\
Input:\\
<English> I ate dinner quickly.\\
<Korean> 나는 저녁을 빨리 먹었다.\\
<Code-Switching> I ate 저녁 빨리.\\
\\
Output:\\
  1. I ate dinner quickly.\\
  2. I ate dinner 빨리.\\
  3. I ate 저녁 빨리.\\
  4. 나는 저녁 빨리 ate.\\
  5. 나는 저녁을 빨리 먹었다.\\
\\
EXAMPLE 2\\
Input:\\
<English> Dad was about to throw away my tooth.\\
<Korean> 아빠가 내 이빨을 빼려고 했어.\\
<Code-Switching> 아빠 was about to 뺄래 my 이빨.\\
\\
Output:\\
  1. Dad was about to throw away my tooth.\\
  2. Dad was about to throw away 내 이빨.\\
  3. 아빠 was about to 뺄래 my 이빨.\\
  4. 아빠 was about to 내 이빨 빼려고 했어.\\
  5. 아빠가 내 이빨을 빼려고 했어.\\
\\
Example 3\\
Input:\\
<English> Tom thinks Bill likes himself.\\
<Korean> 톰은 빌이 자기 자신을 좋아한다고 생각한다.\\
<Code-Switching> Tom thinks that Bill이 자기를 좋아한다.\\
\\
Output:\\
  1. Tom thinks Bill likes himself.\\
  2. Tom thinks Bill likes 자기.\\
  3. Tom thinks that Bill이 자기를 좋아한다.\\
  4. Tom은 Bill이 자기를 좋아한다고 think한다.\\
  5. 톰은 빌이 자기 자신을 좋아한다고 생각한다.\\
\\
{[BEGIN TASK]}
\end{tcolorbox}
    
\begin{tcolorbox}[breakable, enhanced, top=1pt, left=1pt, right=1pt, bottom=1pt, colback=white, title=Prompt for generating gradual code-switching (Ko\textrightarrow En) few-shot demonstrations]
\small
You are a bilingual rewriting assistant.\\
Your task is to generate five versions of a sentence that gradually transition from Korean to English.\\
\\
{[INPUT]}\\
  • One English sentence (E)\\
  • Its Korean translation (K)\\  
  • A code-switching version of the sentence (C), where about 50\% of Korean words are replaced by English equivalents\\
\\
{[OUTPUT]}\\
Generate a sequence of five sentences showing a smooth progression from Korean to English:\\
  1. Korean only (100\% Korean, source syntax S-O-V)\\
  2. 75\% Korean + 25\% English (matrix language: Korean, embedded language: English)\\
  3. 50\% Korean + 50\% English (matrix language: Korean, embedded language: English)\\
  4. 25\% Korean + 75\% English (matrix language: Korean, embedded language: English)\\
  5. English only (100\% English, target syntax S-V-O)\\
\\
{[RULES]}\\
Following the Matrix Language Frame (MLF) model,\\
  • Preserve Korean word order (S-O-V) and syntax until version 5 (full English).\\
  • Use English equivalents from E when inserting English into Korean sentences.\\
  • Keep the code-switching natural and consistent, not random.\\
  • Do not add explanations, notes, or extra text — output only the five sentences in order.\\
\\
{[EXAMPLES]}\\
\\
EXAMPLE 1\\
Input:\\
<Korean> 미나야, 바구니에 장난감을 다 넣고 빨리 집에 가자.\\
<English> Meena, put all the toys in the basket quickly, and go home.\\
<Code-Switching> Meena, basket 안에다 all the toys를 빨리 put하고 집에 가자.\\
\\
Output:\\
  1. 미나야, 바구니에 장난감을 다 넣고 빨리 집에 가자.\\
  2. Meena, 바구니에 장난감을 다 put하고 빨리 집에 가자.\\
  3. Meena, basket 안에다 all the toys를 빨리 put하고 집에 가자.\\
  4. Meena, put all the toys in the basket quickly, 집에 가자.\\
  5. Meena, put all the toys in the basket quickly, and go home.\\
\\
EXAMPLE 2\\
Input:\\
<Korean> 지난 번에 도로교통공단이 내 운전면허증을 실수로 갱신하지 않았어요.\\
<English> Last time, by mistake, they did not renew my driver's license at the DMV.\\
<Code-Switching> 지난번에 motor vehicle department에서 내 driver's license를 mistake로 갱신하지 않았어요.\\
\\
Output:\\
  1. 지난 번에 도로교통공단이 내 운전면허증을 실수로 갱신하지 않았어요.\\
  2. 지난 번에 motor vehicle department가 내 운전면허증을 실수로 갱신하지 않았어요.\\
  3. 지난번에 motor vehicle department에서 내 driver's license를 mistake로 갱신하지 않았어요.\\
  4. Last time, motor vehicle department에서 my driver's license를 mistake로 renew하지 않았어요.\\
  5. Last time, by mistake, they did not renew my driver's license at the DMV.\\
\\
Example 3\\
Input:\\
<Korean> 존은 메리가 어제 무엇을 샀는지 궁금해한다.\\
<English> John wonders what Mary bought yesterday.\\
<Code-Switching> John은 Mary가 yesterday 무엇을 샀는지 궁금해한다.\\
\\
Output:\\
  1. 존은 메리가 어제 무엇을 샀는지 궁금해한다.\\
  2. John은 메리가 yesterday 무엇을 샀는지 궁금해한다.\\
  3. John은 Mary가 yesterday what을 샀는지 궁금해한다.\\
  4. John wonders what Mary가 yesterday 샀는지 궁금해한다.\\
  5. John wonders what Mary bought yesterday.\\
\\
{[BEGIN TASK]}
\end{tcolorbox}

\subsection{Code-switching In-context Learning}
\begin{tcolorbox}[breakable, enhanced, top=1pt, left=1pt, right=1pt, bottom=1pt, colback=white, title=Prompt for \method]
\small
Answer the following questions written in non-English by explicitly showing a step-by-step translation process into English, then provide the final answer.\\
\\
**Core Behaviors**\\
\\
1. Mandatory Opening Self-Instruction\\
  - Always begin with the exact sentence: "Let's gradually translate this non-English query into English, then think in English, and finally answer the question."\\
2. Gradual Code-Switching Output:\\
  - Show the transformation from the original query to English in 5 steps.\\
  - Each step progressively replaces more non-English words with English until the query is fully natural English.\\
  - End with a clean, natural English rendering of the question.\\
3. Answer Format: \\
  - End with the final answer in the exact format: "The answer is X", where `X` is a single uppercase letter of the correct choice (e.g., A, B, C, D).\\
  - No explanations or justifications after the answer.\\
\\
Unlike hidden scratchpad reasoning, note that the translation and reasoning process must be explicitly output. Keep the structure identical across all responses.\\
\\
\texttt{\{Few-shot demonstrations\}}
\end{tcolorbox}

\subsection{Evaluation Experiments}

\begin{tcolorbox}[breakable, enhanced, top=1pt, left=1pt, right=1pt, bottom=1pt, colback=white, title=Prompt for multiple-choice QA experiments]
\small
You will be asked to answer a multiple-choice question. Read the following question and choices then select the single best answer.\\
\\
- Output only the single uppercase letter of the correct choice (e.g., A, B, C, D).\\
- Do not output anything else: no reasoning, no explanation, no restating the question, no prefixes like “Answer:”, no punctuation, spaces, or newlines.\\
\\
The final output must be exactly one letter.
\end{tcolorbox}

\begin{tcolorbox}[breakable, enhanced, top=1pt, left=1pt, right=1pt, bottom=1pt, colback=white, title=Prompt for short-answer QA experiments]
\small
You will be asked to answer a short-answer question. Read the following question and provide a single answer without any explanations.\\
\\
- Output only the single answer.\\
- Do not output anything else: no reasoning, no explanation, no restating the question, no prefixes like “Answer:”.
\end{tcolorbox}

\begin{tcolorbox}[breakable, enhanced, top=1pt, left=1pt, right=1pt, bottom=1pt, colback=white, title=Prompt for machine translation experiments]
\small
You will be asked to translate a \{\texttt{target language}\} text. Read the following sentence and translate it into English without any explanations.\\
\\
- Output only the English equivalent.\\
- Do not output anything else: no reasoning, no explanation, no restating the question, no prefixes like “Answer:”.
\end{tcolorbox}

\clearpage
\section{Additional Related Work}
In this section, we provide further discussions on prior studies related to code-switching.

\paragraph{Code-switching.}
Code-switching, also known as code-mixing or language alternation, is a common linguistic phenomenon in which a speaker interleaves two or more languages within a single conversational context~\citep{auer1998code}.
It occurs in various switching levels: subwords such as at morpheme boundary (\ie intra-word switching), tag phrases (\ie tag-switching), words (\ie intra-sentential switching), and sentences or clauses (\ie inter-sentential switching).
Matrix Language Frame (MLF) model, which distinguishes a grammatically dominant \emph{matrix language} (ML) and an inserted \emph{embedded language} (EL) ~\citep{myers1997duelling}, has been widely adopted as a syntactic rule for code-switching.
The following examples illustrate code-switching, with variation in the choice of matrix language and embedded language. 
The sentence originates from Global MMLU~\citep{singh-etal-2025-global}: ``\emph{Paper will burn at approximately what temperature in Fahrenheit?}''

\lingset{glhangstyle=none, glspace=1em}

\ex
\begingl
\gla Paper will \textbf{brûler} at approximately \textbf{quelle} \textbf{température} in Fahrenheit ? //
\glb ML    ML   EL              ML   ML          EL              EL            ML            ML //
\glft (ML: English, EL: French) //
\endgl
\xe

\ex
\begingl
\gla En degrés Fahrenheit, à \textbf{approximately} \textbf{what} \textbf{temperature} le papier \textbf{burn} -t-il ? //
\glb ML  ML      ML          ML EL          EL                      EL              ML  ML     EL   ML      //
\glft (ML: French, EL: English) //
\endgl
\xe

\paragraph{Code-switching text generation.}
Only a limited number of code-switching corpora and labeled datasets exist for specific language pairs, and code-switching among non-English languages is hardly available~\citep{winata-etal-2023-decades}.
Hence, the synthetic generation of code-switching has become the primary strategy.
Early efforts such as \citet{jayanthi-etal-2021-codemixednlp, rizvi-etal-2021-gcm} proposed toolkits for Hindi-English, although they lack generalizability across languages.
More recently, prompting LLMs (\ie GPT-4) has been widely adopted for synthetic data generation~\cite { yoo-etal-2025-code-switching, yoo-etal-2025-code, kim2025codeswitched, yang2025codemixbench}.
In particular, \citet{yoo-etal-2025-code-switching} further provided quantitative and qualitative analyses comparing LLM-generated code-switching text with human bilingual data, showing that such text largely follows natural switching patterns with one distinctive feature---redundant synonyms appearing in both languages---which may actually help LLMs by providing explicit lexical alignment cues.
Therefore, a slight ``unnaturalness'' of synthetic code-switching may not be a weakness but rather a source of beneficial noise that facilitates multilingual transfer.
In addition, \citet{xie2025switchlingua} introduced LinguaMaster, a LLM multi-agent collaboration framework specifically designed for controlled code-switching generation.

\paragraph{Code-switching for multilingual evaluations.}
As the multilingual capabilities of LLMs advance, recent benchmarks range from broad NLP tasks~\citep{yang2025codemixbench, abdaljalil2025evaluating, mohamed-etal-2026-lost} to domain-specific evaluations, including translation~\citep{huzaifah-etal-2024-evaluating, zhang2025chai}, dialogue summarization~\citep{suresh2025cssum}, and red-teaming~\citep{song-etal-2025-multilingual, yoo-etal-2025-code}.
Despite these efforts, \citet{zhang-etal-2023-multilingual} reported that LLMs still struggle with code-switching, sometimes underperforming compared to smaller, fine-tuned systems. 
This suggests that code-switching remains an unsolved challenge for multilingual evaluation, rather than a solved proxy task.

\section{Additional Results and Analyses}

\subsection{Ablation Study}
\label{sec:appendix_ablation}

\paragraph{Scalability of \method.}
\label{sec:scalability}
\begin{wrapfigure}{r}{0.5\linewidth}
    \vspace{-6mm}
    \centering
    \includegraphics[width=\linewidth]{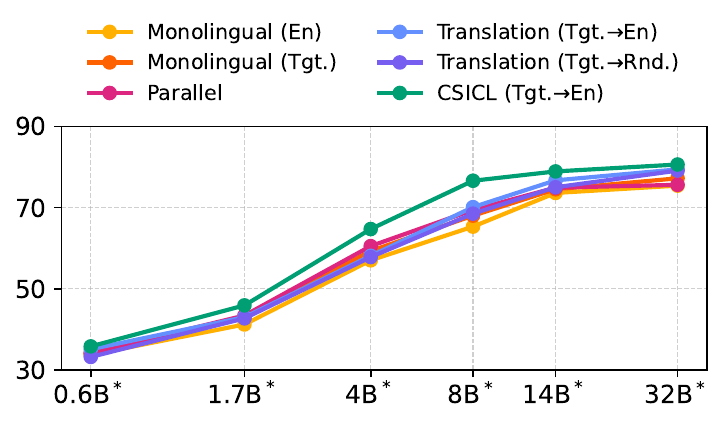}
    \caption{Experimental results of X-ICL approaches in the target language (Ko) on Global MMLU using different sizes of Qwen 3. X-axis is on a log scale.}
    \label{fig:scalability}
    \vspace{-5mm}
\end{wrapfigure}
We examine the effectiveness of \method on small language models (SLMs), which fall behind larger models in instruction-following performance.
We test different sizes of \texttt{Qwen 3} (\ie 0.6B, 1.7B, 4B, 8B, 14B, and 32B) on the sampled set of Global MMLU in Korean as a target language.
Figure~\ref{fig:scalability} reports the target language performance of X-ICL baselines and \method across model sizes, shown in a log scale.
While the gap between \method and X-ICL baselines is particularly pronounced when the model size is larger than 4B, we observe that \method consistently outperforms all X-ICL baselines across model sizes.

\paragraph{Extended Thinking.}

Several frontier models have strong native chain-of-thought or reasoning capabilities.
We investigate whether extended thinking of LLMs diminishes the benefit of \method, by mitigating the latent translation barrier.
Table~\ref{tab:thinking_mode} reports the experimental results of \texttt{Qwen3-32B} on Global MMLU with the thinking mode enabled.
As expected, extended thinking raises the baseline, especially for mid- and low-resource languages, thereby reducing \method’s marginal gains.
However, the gains do not disappear.
\method’s gains slightly reduce from +6.0\textrightarrow +5.8pp in target languages, +1.3\textrightarrow +1.0pp in unseen high-resource languages, +2.1\textrightarrow +1.8pp in unseen mid-resource languages, and +4.8\textrightarrow +2.9pp in unseen low-resource languages.
\method still significantly improves performance in the target and unseen mid- to low-resource languages across all X-ICL baselines.
We further conduct a qualitative error analysis of the model’s reasoning traces.
We observe that the extended thinking often helps the model explicitly translate or paraphrase the non-English query before reasoning.
However, in failure cases, the model sometimes mistranslates a key entity, relation, or constraint early in the trace and then reasons consistently but incorrectly from that corrupted English surrogate~\cite{bafna2025translation}.
Once the reasoning trajectory enters this wrong path, the model rarely self-corrects later.
In contrast, \method often mitigates these failures by making the cross-lingual transition explicit and gradual: key content is preserved across intermediate code-switched steps before the model fully reasons in English. 
This suggests that native extended thinking and \method are partially overlapping but complementary.

\begin{table*}[htb!]
\centering
\resizebox{\linewidth}{!}{
\begin{tabular}{@{}l|llllccccc@{}}
\toprule
\multirow{2}{*}{Method}                                 & \multicolumn{4}{c}{X-ICL setting}                                                                                       & \multirow{2}{*}{En} & \multirow{2}{*}{Tgt.$^*$} & \multicolumn{3}{c}{Unseen Lang.}              \\ \cmidrule(lr){2-5} \cmidrule(l){8-10} 
                                                        & \multicolumn{2}{c|}{Demonstration}               & \multicolumn{2}{c|}{Instruction}                                     &                     &                           & High          & Mid$^*$       & Low$^*$       \\ \midrule
Zero-shot learning                 & \xmark & \multicolumn{1}{l|}{}                   & \xmark & \multicolumn{1}{l|}{}                                       & {\ul 93.3}          & 73.8                      & 88.1          & 67.1          & 47.2          \\ \midrule
\multirow{3}{*}{Few-shot learning} & \cmark & \multicolumn{1}{l|}{Monolingual (En)}   & \xmark & \multicolumn{1}{l|}{}                                       & \textbf{93.4}       & 75.3                      & 88.4          & 67.8          & 49.1          \\
                                   & \cmark & \multicolumn{1}{l|}{Monolingual (Tgt.)} & \xmark & \multicolumn{1}{l|}{}                                       & \textbf{93.1}       & 75.4                      & 86.0          & 67.1          & 47.3          \\ \cmidrule(l){2-10} 
                                   & \cmark & \multicolumn{1}{l|}{Parallel}           & \xmark & \multicolumn{1}{l|}{}                                       & {\ul 93.3}          & 76.1                      & {\ul 88.5}    & 67.7          & 48.8          \\ \midrule
\multirow{2}{*}{Zero-shot CoT}     & \xmark & \multicolumn{1}{l|}{}                   & \cmark & \multicolumn{1}{l|}{Translation (Tgt.\textrightarrow En)}   & \textbf{93.2}       & 76.8                      & 88.4          & {\ul 68.5}    & {\ul 49.5}    \\
                                   & \xmark & \multicolumn{1}{l|}{}                   & \cmark & \multicolumn{1}{l|}{Translation (Tgt.\textrightarrow Rnd.)} & 93.0                & {\ul \textbf{78.8}}       & \textbf{86.2} & 65.2          & 49.4          \\ \midrule
\method                            & \cmark & Gradual CS (Tgt.\textrightarrow En)     & \cmark & Gradual Translation (Tgt.\textrightarrow En)                & 93.2                & \textbf{81.1}             & \textbf{89.4} & \textbf{69.6} & \textbf{52.0} \\ \bottomrule
\end{tabular}
}
\caption{Ablation results comparing \method to existing X-ICL baselines using Qwen 3 with thinking mode enabled. X-ICL setting describes each method in terms of a combination of its few-shot demonstrations and instruction. \method outperforms X-ICL baselines in both target and unseen languages, yielding stable results in English.}
\label{tab:thinking_mode}
\end{table*}

\paragraph{Demonstration Sensitivity.}

We test the sensitivity of the demonstration of \method. 
Table~\ref{tab:demonstration_sensitivity} reports the experimental results on Global MMLU using \texttt{Qwen3-32B} under five different random seeds and dynamic demonstrations, where the target language is Korean.
Specifically, we use the same balanced Global MMLU subset as in the main experiments, consisting of 3.6k instances.
We first convert all Korean instances into candidates for gradual code-switching demonstrations using the \method construction pipeline.
Then, for each evaluation query, we randomly sample 5 gradual code-switching demonstrations from the remaining instances, explicitly excluding the current query itself.
We repeat this entire process with 5 different random seeds.
Under this randomized demonstration-sampling setting, \method achieves 81.5$\pm$1.2 accuracy, while the strongest X-ICL baseline achieves 78.8. The results show that \method is not dependent on the particular demonstrations or seeds.
\method outperforms all X-ICL baselines in the target language and mid- to low-resource unseen languages.

\begin{table}[htb!]
\resizebox{\linewidth}{!}{
\begin{tabular}{@{}l|c|c|cc|cc|cc@{}}
\toprule
X-ICL setting         & En                & Tgt.$^*$          & Zh                & Es                & Id$^*$            & Tr$^*$            & Sw$^*$            & Te$^*$            \\ \midrule
Monolingual (En)      & \textbf{89.1$\pm$0.4} & 75.5$\pm$0.4          & 89.4$\pm$0.2          & 83.8$\pm$0.3          & 65.3$\pm$0.5          & 60.7$\pm$0.2          & 44.2$\pm$0.5          & 37.6$\pm$0.3          \\
Monolingual (Tgt.)    & 88.7$\pm$0.2          & 77.0$\pm$0.3          & 90.0$\pm$0.4          & 83.0$\pm$0.2          & 64.7$\pm$0.2          & {\ul 61.6$\pm$0.5}    & 42.3$\pm$0.3          & 35.8$\pm$0.6          \\ \midrule
Parallel              & 88.6$\pm$0.4          & 75.6$\pm$0.2          & 89.7$\pm$0.4          & 83.3$\pm$0.2          & 65.0$\pm$0.4          & 61.1$\pm$0.4          & 44.6$\pm$0.3          & 37.7$\pm$0.3          \\ \midrule
Translation (Tgt.→En) & {\ul 88.9$\pm$0.6}    & {\ul 78.8$\pm$0.2}    & \textbf{90.1$\pm$0.2} & 84.1$\pm$0.3          & {\ul 65.2$\pm$0.2}    & 61.1$\pm$0.4          & {\ul 44.9$\pm$0.2}    & 38.5$\pm$0.2          \\
Translation (Rnd.→En) & 88.4$\pm$0.4          & 78.6$\pm$0.4          & 89.9$\pm$0.1          & \textbf{84.5$\pm$0.3} & 65.1$\pm$0.5          & 61.0$\pm$0.2          & 44.8$\pm$0.3          & {\ul 38.3$\pm$0.2}    \\ \midrule
\method (Tgt.→En)     & 88.5$\pm$0.5          & \textbf{81.5$\pm$1.2} & \textbf{90.1$\pm$0.8} & {\ul 84.2$\pm$1.0}    & \textbf{67.5$\pm$1.3} & \textbf{63.0$\pm$1.1} & \textbf{48.3$\pm$1.4} & \textbf{41.6$\pm$1.8} \\ \bottomrule
\end{tabular}
}
\caption{Ablation results comparing \method to existing X-ICL baselines using Qwen 3 on Global MMLU with five different random seeds and dynamic demonstrations.}
\label{tab:demonstration_sensitivity}
\end{table}

\subsection{Is \method More Than Translation or Prompt Expansion?}
\label{sec:translation_paraphrasing}

Table~\ref{tab:main_result_others} reports experimental results on \texttt{Qwen3-32B} using other X-ICL baselines, controlling the combinations of few-shot demonstrations setting and instruction setting, where \method outperforms all X-ICL baselines.

\begingroup
\setlength{\tabcolsep}{4pt}
\begin{table*}[h]
\centering
\resizebox{\linewidth}{!}{
\begin{tabular}{@{}l|llll|ccccc@{}}
\toprule
\multirow{2}{*}{Method}            & \multicolumn{4}{c|}{X-ICL setting}                                                                                        & \multirow{2}{*}{En} & \multirow{2}{*}{Tgt.$^*$} & \multicolumn{3}{c}{Unseen Lang.}              \\ \cmidrule(lr){2-5} \cmidrule(l){8-10} 
                                   & \multicolumn{2}{c|}{Demonstration}                                & \multicolumn{2}{c|}{Instruction}                      &                     &                       & High$^*$      & Mid$^*$       & Low$^*$       \\ \midrule
\multirow{2}{*}{Zero-shot CoT}     & \xmark & \multicolumn{1}{l|}{}                                    & \cmark & Translation (Tgt.\textrightarrow En) + Reason in En & {\ul 88.7}          & 74.4                  & 87.5          & 63.7     & 42.3        \\ 
                                & \cmark & \multicolumn{1}{l|}{Monolingual (Tgt.)}                  & \cmark & Translation (Tgt.\textrightarrow Rnd.-High)  & {\ul 88.7}          & 74.0                  & {\ul 87.6}    & 63.7          & 42.1          \\ \midrule
\multirow{4}{*}{Few-shot learning} & \cmark & \multicolumn{1}{l|}{Monolingual (Tgt.)}                  & \cmark & Translation (Tgt.\textrightarrow En)         & \textbf{88.8}       & {\ul 75.3}            & 87.2          & 63.3          & 41.8          \\
                                   & \cmark & \multicolumn{1}{l|}{Parallel}                            & \cmark & Translation (Tgt.\textrightarrow En)         & {\ul 88.7}          & 75.0                  & 87.4          & 64.0     & {\ul 43.2}   \\ \cmidrule(l){2-10}
                                   & \cmark & \multicolumn{1}{l|}{Parallel}                            & \cmark & Translation (Tgt.\textrightarrow En) + Reason in En & {\ul 88.7}          & 75.0                  & 87.4          & {\ul 64.2}     & 43.1    \\ \cmidrule(l){2-10}
                                   & \cmark & \multicolumn{1}{l|}{Monolingual (Tgt.)}                  & \cmark & Translation (Tgt.\textrightarrow Rnd.)       & 88.6                & 74.8                  & 87.5    & 63.5          & 42.0          \\ \midrule
\multirow{1}{*}{\method}           & \cmark & \multicolumn{1}{l|}{Gradual CS (Tgt.\textrightarrow En)} & \cmark & Gradual Translation (Tgt.\textrightarrow En) & 88.6                & \textbf{76.8}         & \textbf{87.8} & \textbf{64.9} & \textbf{46.0} \\ \bottomrule                           
\end{tabular}
}
\caption{Ablation results of other X-ICL baselines using Qwen 3, controlling the few-shot demonstrations setting and the instruction setting.}
\label{tab:main_result_others}
\end{table*}
\endgroup

\paragraph{Translation baselines.}
We add an ablation that explicitly translates the query into English and instructs the model to reason in English.
Translation vs. Translation-then-reason baseline yields comparable results, and both consistently remain below \method.
This suggests that \method is not merely improving translation quality but facilitating cross-lingual representational alignment.

\paragraph{Pivot language.}
We also add an ablation using random high-resource, random pivot languages (\ie German, Russian, Dutch, and Japanese), ensuring they do not overlap with the evaluation languages.
Translation to random vs. random high-resource pivot languages yields comparable results, and both are consistently outperformed by \method. 

\paragraph{Paraphrasing control.}

\begin{wraptable}{R}{0.5\linewidth}
\centering
\resizebox{\linewidth}{!}{
\begin{tabular}{@{}l|ccccc@{}}
\toprule
\multirow{2}{*}{X-ICL setting}      &\multirow{2}{*}{En} & \multirow{2}{*}{Tgt.$^*$} & \multicolumn{3}{c}{Unseen Lang.}              \\ \cmidrule(l){4-6} 
                                    &                    &                       & High$^*$      & Mid$^*$       & Low$^*$       \\ \midrule
Paraphrasing (En)                   & \textbf{88.9}      & 71.0                  & 86.6          & {\ul 63.2}    & {\ul 41.6}    \\
Paraphrasing (Tgt.)                 & 88.8               & {\ul 72.3}            & {\ul 87.0}    & 62.6          & 39.3          \\ \midrule
\method (Tgt.\textrightarrow En)    & {\ul 88.6}         & \textbf{76.8}         & \textbf{87.8} & \textbf{64.9} & \textbf{46.0} \\ \bottomrule
\end{tabular}
}
\caption{Experimental results comparing \method to paraphrased, monolingual demonstrations, keeping the number of sentences per shot.}
\label{tab:paraphrase_result}
\vspace{-3mm}
\end{wraptable}

\method may benefit from using a larger number of sentences, since \method includes four additional sentences per demonstration.
To control for this, we fix the number of sentences per demonstration to five, the same as in \method, and compare it against monolingual demonstrations.
Specifically, for each shot, we add four paraphrased monolingual demonstrations, generated by GPT-5, and then ask the LLMs to solve the task.

Table~\ref{tab:paraphrase_result} describes experimental results of \method and paraphrased demonstrations on Global MMLU, averaged over Qwen 3 and Gemini 2.5. 
Although paraphrasing provides a marginal improvement in cross-lingual transfer for both target (+0.2pp) and unseen languages (+0.3pp) on average, its effect is limited; \method consistently outperforms both baselines (+6.0pp and +4.8pp in the target and the unseen languages, respectively, on average). 
This suggests that the gains of \method cannot be attributed to a larger demonstration budget alone. 
Instead, the gradual transition across languages supplies a distinct cross-lingual signal that paraphrasing fails to capture, reinforcing the role of code-switching in aligning multilingual representations.

We extend this examination by controlling the number of English and Korean few-shot demonstrations.
Table~\ref{tab:paraphrase_ablation} shows experimental results on Global MMLU using \texttt{Qwen 3} under different combinations of paraphrasing baselines.
We observe mixed results across the paraphrasing baselines for target language and for unseen languages in high- and mid-resource languages.
Interestingly, we find that increasing English demonstrations helps only in unseen low-resource languages.
However, \method consistently outperforms all paraphrasing baselines across target and unseen languages.

\begin{table}[htb!]
\small
\centering
\begin{tabular}{@{}cc|ccccc@{}}
\toprule
\multirow{2}{*}{\# En} & \multirow{2}{*}{\# Tgt.} & \multirow{2}{*}{En} & \multirow{2}{*}{Tgt.$^*$} & \multicolumn{3}{c}{Unseen Lang.}        \\ \cmidrule(l){5-7} 
                      &                         &                     &                       & High$^*$      & Mid$^*$       & Low$^*$       \\ \midrule
5                     & 0                       & \textbf{88.9}       & 71.0                  & 86.6          & {\ul 63.2}    & {\ul 41.6}    \\
4                     & 1                       & 88.6                & 71.3                  & 86.8          & {\ul 63.2}    & 41.0          \\
3                     & 2                       & \textbf{88.9}       & 71.7                  & 86.9          & 62.7          & 40.7          \\
2                     & 3                       & 88.7                & 72.0                  & {\ul 87.1}    & 63.0          & 40.2          \\
1                     & 4                       & 88.7                & {\ul 72.5}            & 86.7          & 62.7          & 39.5          \\
0                     & 5                       & 88.8                & 72.3                  & 87.0          & 62.6          & 39.3          \\ \midrule
\multicolumn{2}{l|}{\method}                    & 88.6                & \textbf{76.8}         & \textbf{87.8} & \textbf{64.9} & \textbf{46.0} \\ \bottomrule
\end{tabular}
\caption{Experimental results comparing \method to paraphrasing baselines, controlling the number of English and target language few-shot demonstrations. \# En and \# Tgt. denote the number of English demonstrations and the number of target language demonstrations, respectively.}
\label{tab:paraphrase_ablation}
\end{table}

\subsection{Human Validation on COMET}
\label{sec:human_validation}
We additionally conduct a pairwise human evaluation to validate the experimental results obtained with COMET.
We randomly sample 200 FLORES+ examples and compare \method against the best-performing baseline for each source language (\ie Parallel for target language translation and Translation (Tgt.\textrightarrow Rnd.) for unseen language translation, respectively).
Following the WMT practice, one author, who is a native speaker of Korean and fluent in English, is shown two anonymized translations (\method vs. baseline) and asked to choose ``A wins / B wins / tie'' based on overall adequacy and fluency.
\method is preferred in 63\% of cases (and 16\% ties) on target-language performance, and in 68\% of cases (and 9\% ties) on unseen-language performance, respectively.
These results provide additional evidence that \method produces more adequate and fluent translations, as measured by both automatic metrics and human preference.

\subsection{Full Results}
\label{sec:full_results}

Tables~\ref{tab:full_qwen}, \ref{tab:full_deepseek}, \ref{tab:full_grok}, and \ref{tab:full_gemini} present experimental results of X-ICL approaches on Global MMLU using Qwen 3, DeepSeek 3.1, Grok 4, and Gemini 2.5.
Table~\ref{tab:full_task} presents experimental results of X-ICL approaches on five different tasks using DeepSeek 3.1, Grok 4, and Gemini 2.5.

\begin{table*}[htb!]
\centering
\small
    \begin{subtable}[t]{\linewidth}
    \centering
\begin{tabular}{@{}l|c|c|cc|cc|cc@{}}
\toprule
X-ICL setting         & En            & Tgt.$^*$      & Zh            & Es$^*$        & Id$^*$        & Tr$^*$        & Sw$^*$        & Te$^*$        \\ \midrule
Monolingual (En)      & 88.6          & 83.5          & 89.5          & 83.2          & 64.8          & 60.5          & 44.6          & 38.2          \\
Monolingual (Tgt.)    & \textbf{88.7} & 83.3          & 90.3          & {\ul 83.8}    & 64.3          & 59.9          & 42.8          & 36.9          \\ \midrule
Parallel              & 88.5          & 83.7          & 90.4          & 82.9          & 65.1          & 60.6          & 44.7          & 38.0          \\ \midrule
Translation (Tgt.→En) & \textbf{88.7} & 83.6          & \textbf{90.5} & 83.6          & 65.5          & 61.1          & {\ul 45.3}    & 38.8          \\
Translation (Rnd.→En) & \textbf{88.7} & {\ul 83.8}    & 90.2          & 83.7          & {\ul 66.0}    & {\ul 61.2}    & 45.1          & {\ul 39.0}    \\ \midrule
\method (Tgt.→En)     & 88.5          & \textbf{84.9} & \textbf{90.5} & \textbf{83.9} & \textbf{66.6} & \textbf{61.8} & \textbf{48.4} & \textbf{42.8} \\ \bottomrule
\end{tabular}
\caption{Target: French (\emph{high})}
    \end{subtable}
    \vspace{4mm}
    
    \begin{subtable}[t]{\textwidth}
    \centering
\begin{tabular}{@{}l|c|c|cc|cc|cc@{}}
\toprule
X-ICL setting         & En            & Tgt.$^*$      & Zh            & Es            & Id$^*$        & Tr$^*$        & Sw$^*$        & Te$^*$        \\ \midrule
Monolingual (En)      & \textbf{88.9} & 75.4          & 89.3          & 83.7          & 65.1          & 60.8          & 44.2          & 37.9          \\
Monolingual (Tgt.)    & 88.7          & 77.2          & 90.0          & 83.0          & 64.8          & 61.3          & 42.3          & 35.7          \\ \midrule
Parallel              & \textbf{88.9} & 75.6          & 89.7          & 83.6          & 64.9          & {\ul 61.5}    & 44.4          & 37.6          \\ \midrule
Translation (Tgt.→En) & \textbf{88.9} & {\ul 79.4}    & {\ul 90.2}    & 84.1          & {\ul 65.5}    & 61.4          & 44.8          & {\ul 38.6}    \\
Translation (Rnd.→En) & 88.5          & 79.1          & 90.0          & {\ul 84.3}    & 65.2          & 61.2          & {\ul 45.1}    & 38.3          \\ \midrule
\method (Tgt.→En)     & 88.6          & \textbf{80.6} & \textbf{90.3} & \textbf{84.4} & \textbf{67.4} & \textbf{62.8} & \textbf{48.5} & \textbf{41.7} \\ \bottomrule
\end{tabular}
\caption{Target: Korean (\emph{mid})}
    \end{subtable}
    \vspace{4mm}

    \begin{subtable}[t]{\textwidth}
    \centering
\begin{tabular}{@{}l|c|c|cc|cc|cc@{}}
\toprule
X-ICL setting         & En            & Tgt.$^*$      & Zh$^*$        & Es$^*$        & Id$^*$        & Tr$^*$        & Sw$^*$        & Te$^*$        \\ \midrule
Monolingual (En)      & {\ul 88.9}    & 53.5          & 89.8          & 83.5          & 64.8          & 60.8          & 44.6          & 37.7          \\
Monolingual (Tgt.)    & \textbf{89.0} & 55.5          & 87.9          & {\ul 86.4}    & 63.9          & 58.4          & 43.5          & 31.0          \\ \midrule
Parallel              & 88.7          & 58.8          & 90.4          & 85.6          & 64.2          & 61.7          & 48.8          & 34.9          \\ \midrule
Translation (Tgt.→En) & 88.8          & {\ul 60.5}    & 90.3          & 85.7          & 64.9          & 63.8          & {\ul 50.0}    & 34.5          \\
Translation (Rnd.→En) & 88.6          & 58.5          & {\ul 90.6}    & 86.2          & {\ul 65.0}    & {\ul 64.2}    & 44.9          & {\ul 41.4}    \\ \midrule
\method (Tgt.→En)     & 88.7          & \textbf{64.9} & \textbf{90.8} & \textbf{86.9} & \textbf{66.1} & \textbf{64.7} & \textbf{51.2} & \textbf{43.4} \\ \bottomrule
\end{tabular}
\caption{Target: Yoruba (\emph{low})}
    \end{subtable}  

\caption{Full experimental results comparing \method to X-ICL baselines using \texttt{Qwen3-32B}.}
\label{tab:full_qwen}
\end{table*}
\begin{table*}[htb!]
\centering
\small
    \begin{subtable}[t]{\linewidth}
    \centering
\begin{tabular}{@{}l|c|c|cc|cc|cc@{}}
\toprule
X-ICL setting         & En            & Tgt.$^*$      & Zh            & Es            & Id$^*$        & Tr$^*$        & Sw$^*$        & Te$^*$        \\ \midrule
Monolingual (En)      & \textbf{89.6} & 84.6          & 91.2          & 84.3          & 66.2          & 61.5          & 45.3          & 38.8          \\
Monolingual (Tgt.)    & 89.5          & 84.4          & 91.8          & {\ul 84.9}    & 65.7          & 61.0          & 43.6          & 37.5          \\ \midrule
Parallel              & 89.4          & 84.7          & 92.0          & 84.0          & 66.5          & 61.7          & 45.4          & 38.6          \\ \midrule
Translation (Tgt.→En) & \textbf{89.6} & 84.8          & {\ul 92.1}    & 84.8          & 66.8          & 62.1          & {\ul 46.0}    & 39.2          \\
Translation (Rnd.→En) & \textbf{89.6} & {\ul 85.0}    & 91.9          & {\ul 84.9}    & {\ul 67.2}    & {\ul 62.2}    & 45.8          & {\ul 39.3}    \\ \midrule
\method (Tgt.→En)     & 89.4          & \textbf{86.0} & \textbf{92.3} & \textbf{85.1} & \textbf{68.0} & \textbf{62.9} & \textbf{49.1} & \textbf{43.1} \\ \bottomrule
\end{tabular}
\caption{Target: French (\emph{high})}
    \end{subtable}
    \vspace{4mm}
    
    \begin{subtable}[t]{\textwidth}
    \centering
\begin{tabular}{@{}l|c|c|cc|cc|cc@{}}
\toprule
X-ICL setting         & En            & Tgt.$^*$      & Zh            & Es            & Id$^*$        & Tr$^*$        & Sw$^*$        & Te$^*$        \\ \midrule
Monolingual (En)      & \textbf{89.8} & 77.3          & 91.0          & 84.6          & 66.4          & 61.8          & 45.0          & 38.6          \\
Monolingual (Tgt.)    & 89.6          & 79.1          & 91.6          & 83.9          & 66.0          & 62.3          & 43.2          & 36.5          \\ \midrule
Parallel              & \textbf{89.8} & 77.5          & 91.4          & 84.5          & 66.1          & {\ul 62.5}    & 45.2          & 38.3          \\ \midrule
Translation (Tgt.→En) & \textbf{89.8} & {\ul 81.1}    & {\ul 91.8}    & 85.0          & {\ul 66.7}    & 62.4          & 45.6          & {\ul 39.0}    \\
Translation (Rnd.→En) & 89.4          & 80.8          & 91.5          & {\ul 85.1}    & 66.4          & 62.2          & {\ul 45.9}    & 38.7          \\ \midrule
\method (Tgt.→En)     & 89.5          & \textbf{82.4} & \textbf{92.0} & \textbf{85.3} & \textbf{68.6} & \textbf{63.6} & \textbf{49.2} & \textbf{41.7} \\ \bottomrule
\end{tabular}
\caption{Target: Korean (\emph{mid})}
    \end{subtable}
    \vspace{4mm}

    \begin{subtable}[t]{\textwidth}
    \centering
\begin{tabular}{@{}l|c|c|cc|cc|cc@{}}
\toprule
X-ICL setting         & En            & Tgt.$^*$      & Zh            & Es$^*$        & Id$^*$        & Tr            & Sw$^*$        & Te$^*$        \\ \midrule
Monolingual (En)      & {\ul 89.8}    & 55.3          & 91.5          & 84.7          & 66.3          & 61.9          & 45.2          & 38.2          \\
Monolingual (Tgt.)    & \textbf{89.9} & 57.3          & 89.6          & {\ul 87.4}    & 65.5          & 59.5          & 44.1          & 32.0          \\ \midrule
Parallel              & 89.7          & 60.6          & 92.0          & 86.6          & 65.8          & 62.7          & 49.3          & 35.9          \\ \midrule
Translation (Tgt.→En) & {\ul 89.8}    & {\ul 62.3}    & 91.9          & 86.7          & 66.6          & 64.5          & {\ul 50.5}    & 35.6          \\
Translation (Rnd.→En) & 89.6          & 60.2          & {\ul 92.3}    & 87.2          & {\ul 66.7}    & {\ul 65.0}    & 45.4          & {\ul 42.2}    \\ \midrule
\method (Tgt.→En)     & 89.7          & \textbf{66.7} & \textbf{92.5} & \textbf{87.8} & \textbf{67.7} & \textbf{65.4} & \textbf{51.6} & \textbf{44.2} \\ \bottomrule
\end{tabular}
\caption{Target: Yoruba (\emph{low})}
    \end{subtable}  

\caption{Full experimental results comparing \method to X-ICL baselines using \texttt{DeepSeek-chat-v3.1}.}
\label{tab:full_deepseek}
\end{table*}
\begin{table*}[htb!]
\centering
\small
    \begin{subtable}[t]{\linewidth}
    \centering
\begin{tabular}{@{}l|c|c|cc|cc|cc@{}}
\toprule
X-ICL setting         & En            & Tgt.$^*$      & Zh            & Es            & Id$^*$        & Tr$^*$        & Sw$^*$        & Te$^*$        \\ \midrule
Monolingual (En)      & \textbf{88.3} & 83.0          & 89.8          & 83.6          & 65.0          & 60.3          & 44.0          & 37.5          \\
Monolingual (Tgt.)    & 88.1          & 82.9          & 90.5          & {\ul 84.1}    & 64.6          & 60.6          & 42.2          & 36.2          \\ \midrule
Parallel              & 88.2          & 83.3          & 90.6          & 83.4          & 65.2          & 60.9          & 44.2          & 37.2          \\ \midrule
Translation (Tgt.→En) & \textbf{88.3} & 83.2          & {\ul 90.7}    & 84.0          & 65.6          & {\ul 61.3}    & {\ul 44.7}    & 37.9          \\
Translation (Rnd.→En) & 88.1          & {\ul 83.5}    & 90.4          & {\ul 84.1}    & {\ul 65.7}    & {\ul 61.3}    & 44.6          & {\ul 38.1}    \\ \midrule
\method (Tgt.→En)     & 88.0          & \textbf{84.5} & \textbf{90.8} & \textbf{84.3} & \textbf{66.5} & \textbf{61.9} & \textbf{47.7} & \textbf{41.7} \\ \bottomrule
\end{tabular}
\caption{Target: French (\emph{high})}
    \end{subtable}
    \vspace{4mm}
    
    \begin{subtable}[t]{\textwidth}
    \centering
\begin{tabular}{@{}l|c|c|cc|cc|cc@{}}
\toprule
X-ICL setting         & En            & Tgt.$^*$      & Zh            & Es            & Id$^*$        & Tr$^*$        & Sw$^*$        & Te$^*$        \\ \midrule
Monolingual (En)      & \textbf{88.4} & 74.6          & 89.6          & 83.9          & 65.3          & 60.6          & 43.7          & 37.2          \\
Monolingual (Tgt.)    & 88.2          & 76.4          & 90.2          & 83.2          & 65.0          & 61.1          & 41.9          & 35.1          \\ \midrule
Parallel              & \textbf{88.4} & {\ul 78.6}    & 89.9          & 83.8          & 65.1          & {\ul 61.3}    & 44.0          & 36.9          \\ \midrule
Translation (Tgt.→En) & \textbf{88.4} & 74.8          & {\ul 90.4}    & 84.2          & {\ul 65.7}    & 61.2          & 44.3          & {\ul 37.8}    \\
Translation (Rnd.→En) & 88.1          & 78.3          & 90.1          & {\ul 84.4}    & 65.4          & 61.1          & {\ul 44.6}    & 37.5          \\ \midrule
\method (Tgt.→En)     & 88.2          & \textbf{79.7} & \textbf{90.6} & \textbf{84.6} & \textbf{67.1} & \textbf{62.3} & \textbf{47.8} & \textbf{40.8} \\ \bottomrule
\end{tabular}
\caption{Target: Korean (\emph{mid})}
    \end{subtable}
    \vspace{4mm}

    \begin{subtable}[t]{\textwidth}
    \centering
\begin{tabular}{@{}l|c|c|cc|cc|cc@{}}
\toprule
X-ICL setting         & En            & Tgt.$^*$      & Zh            & Es$^*$        & Id$^*$        & Tr$^*$        & Sw$^*$        & Te$^*$        \\ \midrule
Monolingual (En)      & {\ul 88.4}    & 52.7          & 90.1          & 83.8          & 65.1          & 60.7          & 43.9          & 36.9          \\
Monolingual (Tgt.)    & \textbf{88.5} & 54.6          & 88.2          & 86.7          & 64.2          & 58.2          & 42.8          & 30.2          \\ \midrule
Parallel              & 88.2          & {\ul 59.5}    & 90.7          & 85.9          & 64.7          & 61.4          & 47.9          & 34.3          \\ \midrule
Translation (Tgt.→En) & 88.3          & 57.8          & 90.6          & 86.0          & 65.4          & 63.5          & {\ul 49.1}    & 34.0          \\
Translation (Rnd.→En) & 88.1          & 57.6          & {\ul 90.9}    & {\ul 86.5}    & {\ul 65.5}    & {\ul 63.8}    & 44.0          & {\ul 40.6}    \\ \midrule
\method (Tgt.→En)     & 88.2          & \textbf{63.7} & \textbf{91.0} & \textbf{87.0} & \textbf{66.5} & \textbf{64.2} & \textbf{50.3} & \textbf{42.5} \\ \bottomrule
\end{tabular}
\caption{Target: Yoruba (\emph{low})}
    \end{subtable}  

\caption{Full experimental results comparing \method to X-ICL baselines using \texttt{grok-4-fast}.}
\label{tab:full_grok}
\end{table*}
\begin{table*}[htb!]
\centering
\small
    \begin{subtable}[t]{\linewidth}
    \centering
\begin{tabular}{@{}l|c|c|cc|cc|cc@{}}
\toprule
X-ICL setting         & En            & Tgt.$^*$      & Zh            & Es            & Id$^*$        & Tr$^*$        & Sw$^*$        & Te$^*$        \\ \midrule
Monolingual (En)      & \textbf{90.2} & 85.6          & 91.5          & 85.1          & 67.2          & 62.7          & 46.1          & 39.6          \\
Monolingual (Tgt.)    & 90.0          & 85.4          & 92.1          & 85.7          & 66.7          & 62.3          & 44.3          & 38.3          \\ \midrule
Parallel              & 90.1          & 85.8          & 92.2          & 84.9          & 67.5          & 63.0          & 46.3          & 39.4          \\ \midrule
Translation (Tgt.→En) & \textbf{90.2} & 85.9          & {\ul 92.3}    & 85.6          & 67.9          & 63.4          & {\ul 46.9}    & 40.1          \\
Translation (Rnd.→En) & 90.1          & {\ul 86.1}    & 92.1          & {\ul 85.7}    & {\ul 68.3}    & {\ul 63.5}    & 46.8          & {\ul 40.3}    \\ \midrule
\method (Tgt.→En)     & 90.0          & \textbf{87.3} & \textbf{92.5} & \textbf{85.9} & \textbf{69.2} & \textbf{64.1} & \textbf{49.9} & \textbf{44.3} \\ \bottomrule
\end{tabular}
\caption{Target: French (\emph{high})}
    \end{subtable}
    \vspace{4mm}
    
    \begin{subtable}[t]{\textwidth}
    \centering
\begin{tabular}{@{}l|c|c|cc|cc|cc@{}}
\toprule
X-ICL setting         & En            & Tgt.$^*$      & Zh            & Es$^*$        & Id$^*$        & Tr$^*$        & Sw$^*$        & Te$^*$        \\ \midrule
Monolingual (En)      & \textbf{90.3} & 79.3          & 91.3          & 85.5          & 67.6          & 63.0          & 45.8          & 39.4          \\
Monolingual (Tgt.)    & 90.1          & 81.1          & 91.9          & 84.8          & 67.2          & 63.5          & 44.0          & 37.3          \\ \midrule
Parallel              & \textbf{90.3} & 79.5          & 91.6          & 85.4          & 67.3          & {\ul 63.7}    & 46.0          & 39.1          \\ \midrule
Translation (Tgt.→En) & \textbf{90.3} & {\ul 83.0}    & {\ul 92.0}    & 85.9          & {\ul 67.9}    & 63.6          & 46.4          & {\ul 39.9}    \\
Translation (Rnd.→En) & 90.0          & 82.7          & 91.7          & {\ul 86.1}    & 67.6          & 63.5          & {\ul 46.7}    & 39.6          \\ \midrule
\method (Tgt.→En)     & 90.1          & \textbf{84.3} & \textbf{92.2} & \textbf{86.4} & \textbf{69.7} & \textbf{64.8} & \textbf{50.1} & \textbf{42.8} \\ \bottomrule
\end{tabular}
\caption{Target: Korean (\emph{mid})}
    \end{subtable}
    \vspace{4mm}

    \begin{subtable}[t]{\textwidth}
    \centering
\begin{tabular}{@{}l|c|c|cc|cc|cc@{}}
\toprule
X-ICL setting         & En            & Tgt.$^*$      & Zh            & Es            & Id$^*$        & Tr$^*$        & Sw$^*$        & Te$^*$        \\ \midrule
Monolingual (En)      & {\ul 90.3}    & 56.8          & 91.8          & 85.4          & 67.4          & 63.1          & 46.0          & 39.0          \\
Monolingual (Tgt.)    & \textbf{90.4} & 58.9          & 89.8          & {\ul 88.3}    & 66.6          & 60.8          & 44.9          & 31.7          \\ \midrule
Parallel              & 90.2          & 62.1          & 92.4          & 87.5          & 66.9          & 63.8          & 50.1          & 35.6          \\ \midrule
Translation (Tgt.→En) & {\ul 90.3}    & {\ul 63.9}    & 92.3          & 87.6          & 67.7          & 65.7          & {\ul 51.3}    & 35.3          \\
Translation (Rnd.→En) & 90.1          & 61.8          & {\ul 92.6}    & 88.1          & {\ul 67.8}    & {\ul 66.2}    & 46.3          & {\ul 42.0}    \\ \midrule
\method (Tgt.→En)     & 90.2          & \textbf{66.2} & \textbf{92.8} & \textbf{88.6} & \textbf{68.9} & \textbf{66.6} & \textbf{52.5} & \textbf{44.8} \\ \bottomrule
\end{tabular}
\caption{Target: Yoruba (\emph{low})}
    \end{subtable}  

\caption{Full experimental results comparing \method to X-ICL baselines using \texttt{Gemini 2.5 Flash}.}
\label{tab:full_gemini}
\end{table*}

\begin{table*}[htb!]
\centering
\small
    \begin{subtable}[t]{\linewidth}
    \resizebox{\linewidth}{!}{
\begin{tabular}{@{}l|cc|cccccc|cccccc@{}}
\toprule
\multirow{2}{*}{Task type} & \multicolumn{2}{c|}{\multirow{2}{*}{Translation}} & \multicolumn{6}{c|}{Reasoning-oriented} & \multicolumn{6}{c}{Knowledge-intensive} \\ \cmidrule(l){4-15}
& \multicolumn{2}{c|}{} & \multicolumn{3}{c|}{Domain-specific} & \multicolumn{3}{c|}{Math} & \multicolumn{3}{c|}{Cultural} & \multicolumn{3}{c}{Social Bias} \\ \midrule
X-ICL setting & Tgt.$^*$ & Uns.$^*$ & En & Tgt.$^*$ & \multicolumn{1}{c|}{Uns.$^*$} & En & Tgt.$^*$ & Uns.$^*$ & En & Tgt.$^*$ & \multicolumn{1}{c|}{Uns.$^*$} & En & Tgt.$^*$ & Uns. \\ \midrule
Monolingual (En) & 78.5 & 73.1 & 45.2 & 38.4 & \multicolumn{1}{c|}{{\ul 30.6}} & 58.5 & 52.1 & 50.3 & 84.1 & 78.5 & \multicolumn{1}{c|}{{\ul 74.1}} & \textbf{92.5} & 89.1 & 86.8 \\
Monolingual (Tgt.) & 76.5 & 72.4 & 46.5 & 40.1 & \multicolumn{1}{c|}{28.5} & \textbf{59.2} & 49.5 & 48.8 & 83.8 & 72.1 & \multicolumn{1}{c|}{71.9} & 90.1 & 92.0 & 86.2 \\ \midrule
Parallel & {\ul 80.1} & {\ul 79.0} & \textbf{48.0} & {\ul 41.5} & \multicolumn{1}{c|}{29.1} & 57.5 & 51.0 & 50.1 & \textbf{84.5} & {\ul 80.1} & \multicolumn{1}{c|}{73.0} & 90.5 & 92.1 & 86.5 \\ \midrule
Translation (Tgt.\textrightarrow En) & 78.2 & 75.5 & {\ul 47.5} & 40.5 & \multicolumn{1}{c|}{28.0} & 58.4 & 57.1 & 51.5 & 84.0 & 79.5 & \multicolumn{1}{c|}{72.5} & 91.0 & {\ul 92.5} & 87.0 \\
Translation (Rnd.\textrightarrow En) & 77.4 & 74.8 & 46.4 & 41.0 & \multicolumn{1}{c|}{29.8} & 58.8 & {\ul 58.2} & {\ul 52.0} & {\ul 84.2} & 79.0 & \multicolumn{1}{c|}{73.1} & {\ul 92.4} & 90.1 & {\ul 87.8} \\ \midrule
\method & \textbf{85.5} & \textbf{84.8} & 46.8 & \textbf{43.5} & \multicolumn{1}{c|}{\textbf{36.3}} & {\ul 59.0} & \textbf{60.1} & \textbf{56.2} & 84.0 & \textbf{82.5} & \multicolumn{1}{c|}{\textbf{75.8}} & {\ul 92.4} & \textbf{93.2} & \textbf{88.0} \\ \bottomrule
\end{tabular}
}
    \caption{DeepSeek 3.1}
    \end{subtable}
    \vspace{4mm}
    
    \begin{subtable}[t]{\linewidth}
    \resizebox{\linewidth}{!}{
\begin{tabular}{@{}l|cc|cccccc|cccccc@{}}
\toprule
\multirow{2}{*}{Task type} & \multicolumn{2}{c|}{\multirow{2}{*}{Translation}} & \multicolumn{6}{c|}{Reasoning-oriented} & \multicolumn{6}{c}{Knowledge-intensive} \\ \cmidrule(l){4-15}
& \multicolumn{2}{c|}{} & \multicolumn{3}{c|}{Domain-specific} & \multicolumn{3}{c|}{Math} & \multicolumn{3}{c|}{Cultural} & \multicolumn{3}{c}{Social Bias} \\ \midrule
X-ICL setting & Tgt.$^*$ & Uns.$^*$ & En & Tgt.$^*$ & \multicolumn{1}{c|}{Uns.$^*$} & En & Tgt.$^*$ & Uns.$^*$ & En & Tgt.$^*$ & \multicolumn{1}{c|}{Uns.$^*$} & En & Tgt.$^*$ & Uns.$^*$ \\ \midrule
Monolingual (En) & {\ul 75.2} & 70.8 & 42.4 & 44.8 & \multicolumn{1}{c|}{31.0} & 49.5 & 47.7 & 48.1 & {\ul 82.2} & 77.6 & \multicolumn{1}{c|}{70.3} & {\ul 90.5} & 87.4 & 80.1 \\
Monolingual (Tgt.) & 73.8 & 70.2 & 43.6 & {\ul 45.1} & \multicolumn{1}{c|}{29.3} & \textbf{50.1} & 48.2 & 48.7 & 81.8 & {\ul 78.2} & \multicolumn{1}{c|}{67.8} & 88.2 & \textbf{90.3} & 82.1 \\ \midrule
Parallel & 71.6 & {\ul 76.6} & \textbf{44.8} & 38.6 & \multicolumn{1}{c|}{30.6} & {\ul 49.8} & 48.1 & 46.1 & \textbf{82.3} & 77.7 & \multicolumn{1}{c|}{70.4} & 88.0 & {\ul 90.0} & 81.6 \\ \midrule
Translation (Tgt.\textrightarrow En) & 70.6 & 73.4 & {\ul 44.6} & 38.0 & \multicolumn{1}{c|}{30.9} & 49.2 & {\ul 49.2} & 49.7 & 81.8 & 76.9 & \multicolumn{1}{c|}{{\ul 70.9}} & \textbf{90.6} & 88.6 & 80.4 \\
Translation (Rnd.\textrightarrow En) & 69.8 & 73.0 & 44.2 & 38.4 & \multicolumn{1}{c|}{{\ul 31.2}} & 48.7 & 48.8 & {\ul 49.8} & 81.3 & 76.5 & \multicolumn{1}{c|}{70.5} & \textbf{90.6} & 88.3 & {\ul 82.9} \\ \midrule
\method & \textbf{76.8} & \textbf{81.6} & 42.8 & \textbf{48.9} & \multicolumn{1}{c|}{\textbf{35.2}} & \textbf{50.1} & \textbf{52.8} & \textbf{50.2} & 82.0 & \textbf{80.8} & \multicolumn{1}{c|}{\textbf{72.1}} & 89.0 & \textbf{90.3} & \textbf{83.8} \\ \bottomrule
\end{tabular}
}
    \caption{Grok 4}
    \end{subtable}
    \vspace{4mm}

    \begin{subtable}[t]{\linewidth}
    \resizebox{\linewidth}{!}{
\begin{tabular}{@{}l|cc|cccccc|cccccc@{}}
\toprule
\multirow{2}{*}{Task type} & \multicolumn{2}{c|}{\multirow{2}{*}{Translation}} & \multicolumn{6}{c|}{Reasoning-oriented} & \multicolumn{6}{c}{Knowledge-intensive} \\ \cmidrule(l){4-15} & \multicolumn{2}{c|}{} & \multicolumn{3}{c|}{Domain-specific} & \multicolumn{3}{c|}{Math} & \multicolumn{3}{c|}{Cultural} & \multicolumn{3}{c}{Social Bias} \\ \midrule
X-ICL setting & Tgt.$^*$ & Uns.$^*$ & En & Tgt.$^*$ & \multicolumn{1}{c|}{Uns.$^*$} & En & Tgt.$^*$ & Uns.$^*$ & En & Tgt.$^*$ & \multicolumn{1}{c|}{Uns.$^*$} & En & Tgt.$^*$ & Uns. \\ \midrule
Monolingual (En) & {\ul 79.5} & 75.2 & 48.6 & 49.2 & \multicolumn{1}{c|}{34.3} & 54.1 & 52.0 & 52.4 & {\ul 86.3} & 85.7 & \multicolumn{1}{c|}{80.5} & 94.6 & 91.5 & 94.2 \\
Monolingual (Tgt.) & 77.8 & 74.5 & 48.8 & 49.6 & \multicolumn{1}{c|}{32.6} & \textbf{54.8} & 52.6 & 53.2 & 85.9 & 84.5 & \multicolumn{1}{c|}{{\ul 81.1}} & 92.5 & {\ul 94.5} & 93.8 \\ \midrule
Parallel & 76.1 & {\ul 81.1} & {\ul 49.1} & {\ul 53.0} & \multicolumn{1}{c|}{35.0} & 54.3 & 54.5 & 50.4 & \textbf{86.5} & 85.8 & \multicolumn{1}{c|}{79.8} & 92.2 & 94.1 & 92.7 \\ \midrule
Translation (Tgt.\textrightarrow En) & 75.3 & 77.8 & 48.8 & 52.4 & \multicolumn{1}{c|}{35.5} & 53.6 & {\ul 55.8} & 54.1 & 85.9 & 86.1 & \multicolumn{1}{c|}{80.2} & {\ul 94.7} & 92.7 & \textbf{94.5} \\
Translation (Rnd.\textrightarrow En) & 74.5 & 77.4 & 48.4 & 52.8 & \multicolumn{1}{c|}{{\ul 36.9}} & 53.1 & 53.2 & {\ul 54.2} & 85.4 & {\ul 86.6} & \multicolumn{1}{c|}{79.8} & \textbf{94.8} & 92.4 & 92.0 \\ \midrule
\method & \textbf{86.2} & \textbf{83.3} & \textbf{49.3} & \textbf{56.5} & \multicolumn{1}{c|}{\textbf{39.8}} & {\ul 54.5} & \textbf{57.6} & \textbf{55.8} & 86.2 & \textbf{88.0} & \multicolumn{1}{c|}{\textbf{82.2}} & 93.3 & \textbf{94.7} & {\ul 94.3} \\ \bottomrule
\end{tabular}
}
    \caption{Gemini 2.5}
    \end{subtable}

\caption{Full experimental results on five diverse tasks and domain knowledge.}
\label{tab:full_task}
\end{table*}

\end{document}